\newcommand{\caesar}{\textsc{caesar}}
\newcommand{\scorpio}{\textsc{Scorpio}}

\documentclass[sn-mathphys]{sn-jnl}%
\usepackage{lineno}
\usepackage{float}
\usepackage{graphicx}
\usepackage{caption}
\usepackage{subcaption}
\usepackage{booktabs}
\usepackage{multirow}
\usepackage{tabularx}
\usepackage{xcolor}
\usepackage{soul}
\usepackage[normalem]{ulem} 
\usepackage{appendix}
\usepackage{hyperref}
\usepackage{adjustbox}
\usepackage{diagbox}
\newcommand{\changed}[1]{\textcolor{black}{#1}}

\jyear{2022}%

\theoremstyle{thmstyleone}%

\theoremstyle{thmstyletwo}%

\theoremstyle{thmstylethree}%

\begin{document}

\title[Radio astronomical images object detection and segmentation]{Radio astronomical images object detection and segmentation: A benchmark on deep learning methods}

\author*[2,1]{\fnm{Renato} \sur{Sortino}}\email{renato.sortino@inaf.it}
\author[3,1]{\fnm{Daniel} \sur{Magro}}
\author[4]{\fnm{Giuseppe} \sur{Fiameni}}
\author[1]{\fnm{Eva} \sur{Sciacca}}
\author[1]{\fnm{Simone} \sur{Riggi}}
\author[3]{\fnm{Andrea} \sur{DeMarco}}
\author[2]{\fnm{Concetto} \sur{Spampinato}}
\author[5]{\fnm{Andrew M.} \sur{Hopkins}}
\author[1]{\fnm{Filomena} \sur{Bufano}}
\author[1]{\fnm{Francesco} \sur{Schillirò}}
\author[1]{\fnm{Cristobal} \sur{Bordiu}}
\author[1,2]{\fnm{Carmelo} \sur{Pino}}
\affil*[1]{\orgdiv{Osservatorio Astrofisico di Catania}, \orgname{INAF}, \orgaddress{Via Santa Sofia 78} \city{Catania}, \postcode{95123} \country{Italy}}
\affil*[2]{\orgdiv{Department of Electrical, Electronic and Computer Engineering}, \orgname{University of Catania}, \city{Catania}, \country{Italy}}
\affil[3]{\orgdiv{Institute of Space Sciences and Astronomy}, \orgname{University of Malta}, \city{Msida}, \postcode{MSD2080} \country{Malta}}
\affil[4]{\orgdiv{NVIDIA AI Technology Centre}, \country{Italy}}
\affil[5]{\orgdiv{Australian Astronomical Optics}, \orgname{Macquarie University}, \orgaddress{105 Delhi Rd} \city{North Ryde}, \postcode{NSW 2113} \country{Australia}}

\label{sec:abstract}
\abstract{In recent years, deep learning has been successfully applied in various scientific domains. Following these promising results and performances, it has recently also started being evaluated in the domain of radio astronomy. In particular, since radio astronomy is entering the Big Data era, with the advent of the largest telescope in the world - the Square Kilometre Array (SKA), the task of automatic object detection and instance segmentation is crucial for source finding and analysis. 
In this work, we explore \changed{the} performance of the most affirmed deep learning approaches, applied to astronomical images obtained by radio interferometric instrumentation, to solve the task of automatic source detection. This is carried out by applying models designed to accomplish two different \changed{kinds} of tasks: object detection and semantic segmentation. The goal is to provide an overview of existing techniques, in terms of prediction performance and computational efficiency, to scientists in the astrophysics community who would like to employ machine learning in their research.}


\maketitle
\section{Introduction} %
\label{sec:introduction}
In recent years, technological advancement in astronomy and astrophysics has marked the need for innovative tools and techniques to process the huge amount of data and images captured from different instruments for radio astronomy observations~\cite{Magro2021,Ralph2019}.
The several types of collected data (images, radio signals, etc.) need to be processed according to the specific tasks for extracting and evaluating useful information to support scientific research.
In particular, in the context of radio-astronomical surveys, the task of object detection and instance segmentation is crucial for extracting information from images to support astrophysics research to catalog and identify the contained objects~\cite{Karypidou2021}.

In contrast to optical instruments, which capture images of the sky's brightness distribution directly, radio interferometers employ interferometry to calculate the two-dimensional discrete intensity distribution of the sky, known as visibility data. A Fourier transform of the visibility data is then performed to produce an image of the sky. The result of this process is the convolution of the true sky brightness with the point spread function (PSF) of the interferometric array, commonly referred to as the dirty image. Due to the incomplete sampling of the interferometric visibility data, the PSF has strong spurious sources that affect the entire image. This can make it difficult to recover the true sky's brightness distribution from interferometric data~\cite{Connor2022}.

Various approaches have been proposed to identify and extract visual information from images, but the majority of these methods are based on classical image processing techniques that show several limitations in terms of accuracy and classification and for other tasks that involve specific features post-analysis (e.g. for identifying extended sources or sources with a complex multi-component morphology).

To overcome these limitations, deep learning models represent the evolution of such approaches and yield interesting results extensively explored in several domains. In particular, object detection and semantic segmentation models based on deep learning are currently used in different domains, such as automotive~\cite{automotive1, automotive2, automotive3, automotive4}, medical imaging~\cite{medical1, medical2, panknet}, video surveillance~\cite{surveillance1, surveillance2}, and robot navigation~\cite{navigation1,navigation2,navigation3}. The radio-astronomical domain has not yet been exhaustively explored with the application of the mentioned methods; therefore, this work represents 
\changed{an attempt to gather performance and computational requirements about several state-of-the-art approaches to be used as a reference for future work.}

In this work, we propose a benchmark to evaluate and compare the performance of multiple object detection and semantic segmentation models based on deep learning (e.g. Mask-RCNN, U-Net, Tiramisu, etc.). We apply these models to astronomical radio images collected from several surveys to detect and classify sources and provide a comprehensive overview of these approaches.

We have performed tests on a dataset consisting of over $10,000$ images containing objects belonging to one of three classes (compact, extended, and spurious sources), extracted from different radio-astronomical surveys images taken with SKA precursors/pathfinders: the Australian Telescope Compact Array (ATCA), the Australian Square Kilometre Array Pathfinder (ASKAP) and the Very Large Array (VLA).

For each model, we evaluated the performance by calculating the F1 score, reliability (precision), and completeness (recall) for each detection. We also explored the performance of subsets of our dataset according to the signal-to-noise ratio (SNR) i.e. the ratio between the peak luminous flux of the objects and the noise component of the image. This allows us to evaluate the detection abilities of the model on faint sources and with a varying degree of noise in the image. 

The analysis conducted in this work shows state-of-the-art object detection and segmentation models applied to radio-astronomical images, providing a baseline for future work.

\section{Related Works} %
\label{sec:related}
The earliest source detection technique was the visual inspection manually carried out by trained astronomers. Needless to say, with the sheer scale and volume of data from modern-day telescopes and surveys, such an approach is intractable and infeasible, as there is far too much data to be manually looked at by astronomers.

Algorithmic techniques represent the first form of automating the source finding task and include a variety of methods based on thresholding and peak detection. To cite a few, Duchamp~\cite{duchamp} allows user-controllable preprocessing followed by a threshold; AEGEAN~\cite{aegean} exploits spatial correlation on data to fit a predictive model; PySE~\cite{pyse} selects islands of high pixel values after removing the background noise, then deblends these islands before fitting a 2D Gaussian model; PyBDSF~\cite{pybdsf} gathers several image decomposition techniques and offers tools to compute the properties of extracted sources.

The natural evolution from this was the use of classic computer vision techniques. For instance,~\cite{friedlander2012latent} applies Latent Dirichlet allocation, a generative statistical model, to image pixels, thus clustering them into either background or source pixels.~\cite{riggi2019caesar} developed another similar technique, which instead performs the source segmentation (or at a lower level, clustering), using the k-means and Self Organizing Maps (SOM)~\cite{schilliro2020SOM} algorithms based on pixels' spatial and intensity values. While these techniques sometimes obtain good results, they are not as capable of generalizing on unseen data as deep learning models, which explains the shift of more recent works towards deep learning techniques for automated source detection.

ConvoSource~\cite{lukic2020convosource} is one such deep learning technique, which uses a relatively lightweight CNN, made up of 3 convolutional layers, dropout, and a fully connected (dense) layer to generate the final output: a binary mask. This output, of course, cannot differentiate between different classes, as it is a binary mask, and thus only performs binary classification.
DeepSource~\cite{vafaei2019deepsource} is another CNN-based model, made up of 5 convolutional layers, with ReLU activations, residual connections, and batch normalization. This model differs from ConvoSource in that the earlier layers are used to boost the signal-to-noise ratio of the input, effectively improving the quality of the image, with a post-processing technique responsible for identifying the predicted sources.
These models use rather simplistic CNNs, and thus the models will not be as capable of recognizing high-level features as state-of-the-art object detection techniques, leading to less than satisfactory performance on more complex or fainter objects.
CLARAN~\cite{wu2019radio} is based on the Faster R-CNN  object detector~\cite{fasterrcnn}, fine-tuned from weights trained on the ImageNet dataset~\cite{russakovsky2015imagenet}, with some architecture changes, such as the RoI Pooling layer replaced by differentiable affine transformations.
Astro R-CNN~\cite{burke2019deblending} applies Mask R-CNN~\cite{he2017mask}, an instance segmentation technique, the evolution of Faster R-CNN, to perform object detection on a simulated dataset.

Mask Galaxy~\cite{farias2020mask} is yet another implementation that uses Mask R-CNN. It fine-tunes a model trained on the COCO dataset~\cite{coco} with astronomical data. This model, however, was only trained to detect one class.

\changed{HeTu~\cite{hetu} uses a combination of residual blocks~\cite{resnet} and a Feature Pyramid Network (FPN) to locate objects in radio images and classify them among four categories} 

While these works prove that great strides have been made in the development of automated source finders, there is still a large margin for improvement. 
\changed{Most of these works, except for HeTu and CLARAN, are trained on simulated datasets, which can limit the capability of these models to generalize to actual telescope data as they can inherit the bias of the acquisition instrument.}

\changed{Object detection methods in the radio astronomical field employ especially architectures based on Mask R-CNN~\cite{farias2020mask,burke2019deblending,maskrcnn-simone} and similar convolutional architectures, using FPNs and ResNets~\cite{wu2019radio,hetu}. In recent years, in the computer vision field, several approaches have pushed forward the state of the art of object detection methods, either improving the existing architectures~\cite{yolov4,yolov7,detectron2,effdet} or employing the transformer architecture, introduced first in the NLP field~\cite{transformers} and then adapted to the vision domain~\cite{vit,detr,deit}. 
One of the families of architectures widely employed as object detectors is the one based on YOLO~\cite{redmon2016you}. YOLOv4~\cite{yolov4} and YOLOv7~\cite{yolov7} present similar architectures, with the latter being the improvement of the former in terms of both performance and efficiency. These models are fully convolutional networks based on FPNs with added modules to improve performance and model size. Similarly, the EfficientDet~\cite{effdet} models build on the EfficientNet~\cite{effnet} model to adapt a well-established convolutional architecture to the object detection task. Transformers introduce the multi-head attention mechanism, which allows the processing of long sequences and has been successfully applied to the vision domain by treating images as sequences of patches and applying the attention mechanism to such sequences. DETR~\cite{detr} is one of the first methods to have applied transformers to object detection by treating the task as a bipartite set matching one, where they match a fixed set of object queries to the detected objects in the image. YOLOS~\cite{yolos} simplifies DETR's architecture by removing the decoder and treating the image patches and the object queries as a single sequence, resulting in a more lightweight model.
A task similar to object detection is semantic segmentation which, instead of predicting bounding boxes, estimates a segmentation mask classifying each pixel of the image. Typical semantic segmentation approaches make use of a U-Net~\cite{ronneberger2015unet} architecture, made of a downsampling encoder, a bottleneck, and an upsampling decoder. The downsampling and upsampling paths are connected by skip connections that avoid gradients from becoming too small and help to keep low-resolution features. Improvements, such as the deep supervision mechanism~\cite{unetds}, contribute to making these models more robust by learning more informative features using an objective function on the hidden layers of the upsampling path. U-Net++~\cite{unetpp} proposes a combination of architectural improvements by improving the skip connection pathways and extending the deep supervision mechanism, while Tiramisu~\cite{tiramisu} employs DenseNets~\cite{densenet} instead of ResNets~\cite{resnet} improving the performance using fewer parameters. Other approaches~\cite{hier-dec,panknet} propose hierarchical decoding for segmentation, which, instead of employing the pipeline of downsampling and upsampling path, decode features at multiple resolutions and combine them at the end.
}

To the best of our knowledge, no study provides an overview of deep learning models, especially relative to semantic segmentation ones, applied to radio-astronomical images. The only work exploring semantic segmentation in radio-astronomy is carried out in~\cite{rg-tiramisu}.
Thus, one of the main contributions of this work will be the standardized comparison of a significant number of state-of-the-art detection and segmentation approaches on a dataset composed of real images from several telescopes. Additionally, this work should serve as a baseline of object detection and segmentation approaches for the radio-astronomical community to orient any scientist who recognizes that deep learning architectures may suit their case study.
A complete summary of other comparable approaches can be found in~\cite{askap-dc,ska-dc}.

\section{\changed{Dataset: Radio Astronomical Images}}
\label{sec:data}
To train and validate the models we made use of a dataset containing $10952$ image cutouts extracted from different radio astronomical survey images taken with the Australian Telescope Compact Array (ATCA), the Australian Square Kilometer Array Pathfinder (ASKAP), and the Very Large Array (VLA) complemented with radio galaxies coming from the Radio Galaxy Zoo (RGZ) project \citep{Banfield2015}.  

Each raw data sample from the surveys comes in a large file size ($\sim4$GB), which is intractable by deep learning models as it would require an excessive amount of resources, so we extract cutouts from each sample. Each cutout may contain multiple objects of the following three classes:
\begin{itemize}
    \item \emph{Extended Sources}: Radio galaxies were taken from the Data Release 1 (DR1)\footnote{\url{https://cloudstor.aarnet.edu.au/plus/s/agKNekOJK87hOh0} from \url{https://github.com/chenwuperth/rgz_rcnn/issues/10}} (Wong et al., in preparation) of the Radio Galaxy Zoo (RGZ) project \citep{Banfield2015}, using 1.4 GHz radio observations at 5" resolution from the Faint Images of the Radio Sky at Twenty cm (FIRST) survey \citep{becker1995first}. The data samples consist of extended sources with a 2- and 3-component morphology. Another sample of such sources was extracted from the 1.2 GHz ASKAP-36 \scorpio{} survey at $\sim$9.4"$\times$7.7" resolution (see~\cite{umana2021first} for a description of the survey). This category includes both extended emission sources and multi-island sources with two, three, or more components. All the sources are likely extragalactic, since HII regions, planetary nebulae, and supernova remnants have been removed.
    \item \emph{Compact sources}: Compact radio sources with single island morphology were taken from the ASKAP-15 and ATCA \scorpio{} surveys, reported above.
    \item \emph{Imaging artifacts}: A collection of imaging artifacts around bright radio sources was obtained from different radio observations: ASKAP EMU pilot survey at $\sim$12.5"$\times$10.9" resolution \citep{Norris2021}, 912 MHz ASKAP-15 \scorpio{} survey at $\sim$24"$\times$21" resolution and the aforementioned 1.2 GHz ASKAP-36 \scorpio{} survey, 2.1 GHz ATCA \scorpio{} survey at $\sim$9.8"$\times$5.8" resolution \citep{Umana2015}. Traditional algorithms extract these artifacts as real sources while they are spurious.
\end{itemize}

\begin{figure}[h]
    \centering
    \includegraphics[width=.7\linewidth]{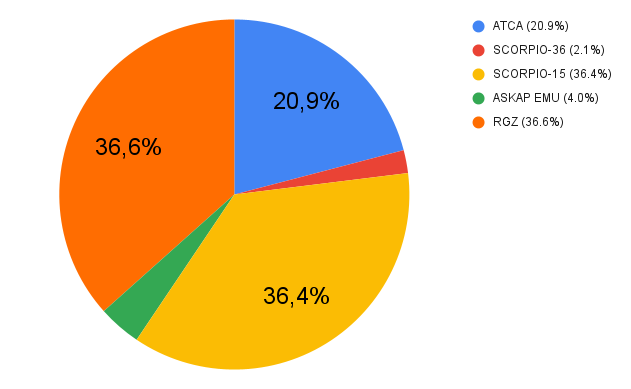}
    \caption{Pie chart of the origin of the images in our dataset.}
     \label{fig:surveys}
\end{figure}

We present a pie chart of the surveys we used to compose our dataset in Figure~\ref{fig:surveys}.
Throughout the paper, we will omit 'source' when it becomes redundant and only use their categorization, i.e. one among \emph{`compact'}, \emph{`extended'}, and \emph{`spurious'}.

The whole dataset, aggregating images from the aforementioned surveys, consists of a total of $36,398$ objects, then split into 3 subsets: training $(70\%)$, validation $(10\%)$ and test $(20\%)$ as shown in Table~\ref{tab:data_split}.
\changed{We use a dataset composed of images acquired using different telescopes, which can help the models better generalize over different image sources. Using images from only one survey can result in a higher bias induced in the network, as analyzed in~\cite{maskrcnn-simone}, Section 4.2.1. We perform a similar analysis on the impact of training on a subset of the data, split by telescope, and report the results in Section~\ref{sec:subsets}}

\begin{table}
    \centering
    \caption{Number of object samples and images for each subset.}
    \begin{tabular}{|c|c|c|c|c|}
        \hline
         \textbf{Object category} & \textbf{Train} & \textbf{Valid.} & \textbf{Test} & \textbf{Total}  \\
         \hline
         Spurious Source & 934 & 133 &	267 & 1,334  \\
         Compact Source	 & 20,603 & 2,943 & 5,886 & 29,432 \\
         Extended Source & 3,940 & 562 &  1,125 & 5,627 \\
          \hline
         \textit{\# of images} & 7,653 & 1,064 & 2,235 & 10,952 \\
         \hline
         \textit{\# object per image (avg.)} & 3.32 & 3.34 & 3.26 & 3.30 \\
          
         \hline
    \end{tabular}
    \label{tab:data_split}
\end{table}

We extracted image cutouts (single-channel, 132$\times$132 pixels, FITS format) from the reference data using the \caesar{} tool~\cite{riggi2019caesar}. The RGZ dataset already provides source cutouts in the same format and bounding boxes for extended source objects. 
We use a dataset composed of image cutouts to train and test the models, as training on the images at the original size would be computationally infeasible. The models we chose for our analysis support image sizes up to $1333 \times 1333$ pixels. Moreover, we resize the images before feeding them to the model.

\begin{figure}[ht]
    \centering
    \begin{subfigure}[b]{.5\textwidth}
      \centering
      \includegraphics[width=.7\linewidth]{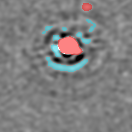}
      \label{fig:sub1}
    \end{subfigure}%
    \begin{subfigure}[b]{.5\textwidth}
      \centering
      \includegraphics[width=.7\linewidth]{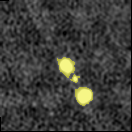}
      \label{fig:sub2}
    \end{subfigure}
    \caption{Image and annotation sample pairs used in training. (a) Compact sources (in red) and spurious sources (in blue). (b) Extended source sample (in yellow)}
    \label{fig:samples}
\end{figure}

While bounding boxes are sufficient for training object detection models, we needed to use segmentation masks to supervise the training of the semantic segmentation models.
For this reason, a raw object segmentation was preliminarily produced with the \caesar{} source finder and later refined by visual examination, needed in particular for spurious sources and bright sources near spurious ones, as source finders often detect them as belonging to the same island.
Images belonging to the RGZ survey have been annotated using both infrared and radio data, while annotations of data originating from other surveys are based only on radio data.
Samples of annotated images are reported in Fig.~\ref{fig:samples}

Raw and processed data are kept under version control, using the Data Version Control (\textsc{dvc}) framework\footnote{\url{https://dvc.org/}}.

\subsection{Data Format} %
\label{sec:data_format}
Input data and annotations represent a crucial element in any machine-learning-based algorithm, as this strongly influences the training procedure, as well as the final performance of the model.
After collecting the images and information about segmentation masks, we need to organize our information so that it can be managed by deep learning models. We introduce a data format to be shared among the models, to avoid the need to adapt the data loading pipeline each time we need to train a new model.  

In our case study, radio-astronomical data and annotations are stored separately in two different formats. The images come as FITS \footnote{\url{https://fits.gsfc.nasa.gov/fits_documentation.html}} files, while the annotations are stored in JSON files, one for each image, where all the individual FITS files containing each object mask are specified. 
Before being fed to any model, images need to go through a pre-processing phase. 
We first remove any inconsistent values (i.e. NaN values) from the FITS data by setting them to the minimum pixel value in the image matrix. Then, we apply a Z-Scale \cite{zwitter2000introduction} normalization with a contrast value of $0.3$ on each image to improve the object visibility before converting the FITS data to PNG. Finally, we convert the data into a PNG image by rescaling all values to the $[0,255]$ range and replicating the single-channel image on three channels. Such transformation is necessary as both object detection and semantic segmentation models are designed to operate on RGB image data.
We did not apply any background subtraction step to the resulting images, as the network is expected to learn the noise pattern from the data.
We acknowledge that these steps can introduce some degree of information loss, as the range of possible values is reduced, and fine-grained details may be lost. It is in our interest to find a better way to preprocess the data, modifying the architecture of some models to exploit the full value range of the images in FITS format, skipping the conversion to PNG.

During training, we employ image augmentation to prevent the model from overfitting. We apply a random number of augmentations (0 to 2) to each image before being fed to the model during training. The available augmentation operations are: (1) \emph{random horizontal flip} (the image is flipped along the x-axis with probability $p$); (2) \emph{$90\deg$ rotation}; and (3) \emph{random resize and cropping} (after resizing the image to a set of possible fixed values, a random area, with a fixed size, is cropped from the image). The number and type of operations applied to a particular image are random and different for each epoch. To avoid confusion, in this work, when we mention epochs, we refer to machine learning training epochs. While training a model, we say it completes an epoch when the whole training dataset has been used to update the model parameters. Training a model requires multiple epochs, meaning that the model will "see" the same dataset multiple times.

\begin{figure}[ht!]
	\centering-
	\includegraphics[width=1\textwidth]{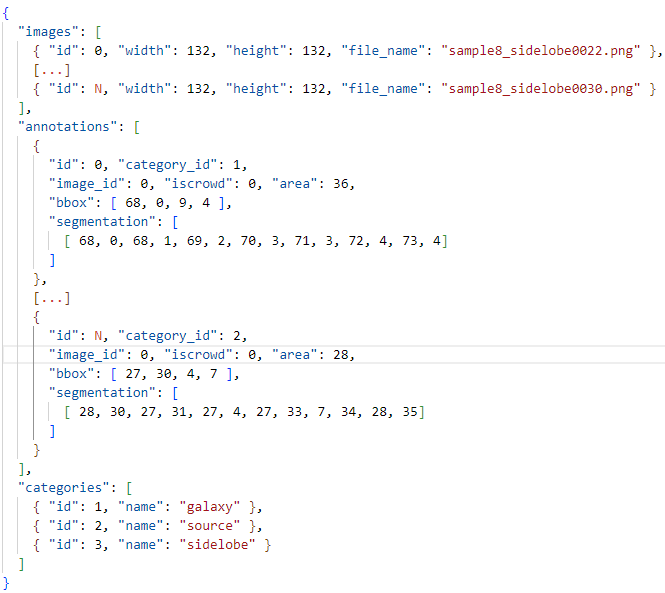}
	\caption{An example of annotation file in JSON (COCO-like) format.}
	\label{fig:json_format}
\end{figure}

For each image, we extract information about the segmentation mask of each object and compute the maximum and minimum of the segmentation masks along the 2D coordinates (x,y) to get the coordinates of the respective bounding box coordinates. Finally, we aggregate the bounding box and segmentation mask information into a single JSON object and we link it to the corresponding image by adding, to each object in the image, the "image id" field. 
In addition to bounding box and segmentation mask annotations, we add to each object description other fields, namely: \textit{category id} to define the type of object for each mask; \textit{area} that indicates the extension covered by the object, in terms of original pixel size, trivially calculated by multiplying the width and height of the bounding boxes; \textit{iscrowd} that specifies if a region contains a set of packed objects of the same class.
We add these attributes to conform to the COCO \cite{coco} format, as many object detection models, including the ones we tested, rely on this kind of annotation. 
A visualization of a final JSON annotation file is shown in Fig.~\ref{fig:json_format}.

\section{Architectures} 
\label{sec:methodology}
We investigate several deep learning models, performing object detection and semantic segmentation, to identify and classify sources. We mainly explore architectures based on CNN and Transformers~\cite{transformers}. Most CNN-based object detection models rely on Region Proposal Networks~\cite{fasterrcnn}, which are computationally expensive modules that propose several regions that could potentially contain objects and then select only a few of them. This technique is more efficient than the selective search~\cite{selective-search} algorithm used in previous approaches. Single-stage detection algorithms~\cite{ssd} are CNNs that avoid the use of RPNs by dividing the image on a grid and assigning a confidence score to one or more proposed predictions within each grid cell. Transformers are a more recent family of architectures, originally designed for Natural Language Processing tasks which have been extended to the vision domain~\cite{vit}. This last work paved the road for a whole family of models, which have been applied to several tasks in the vision domain, such as image classification~\cite{crossvit, cait, sst}, object detection~\cite{detr,yolos,ddetr,tspdet}, and knowledge distillation~\cite{deit, minilm}.

Semantic segmentation models follow a different architecture. They are characterized by encoder-decoder pipelines, where images are first encoded to a low-dimensionality representation by downsampling the feature maps at each encoding stage to generate its latent representation. Then the image is expanded back to the original image size in the decoding path to obtain the binary masks that classify the objects. There are many variants that extend this framework by applying skip connections~\cite{ronneberger2015unet}, deep supervision~\cite{unetds}, or hierarchical decoding~\cite{panknet}.

\begin{figure*}
	\centering
	\includegraphics[width=1\textwidth]{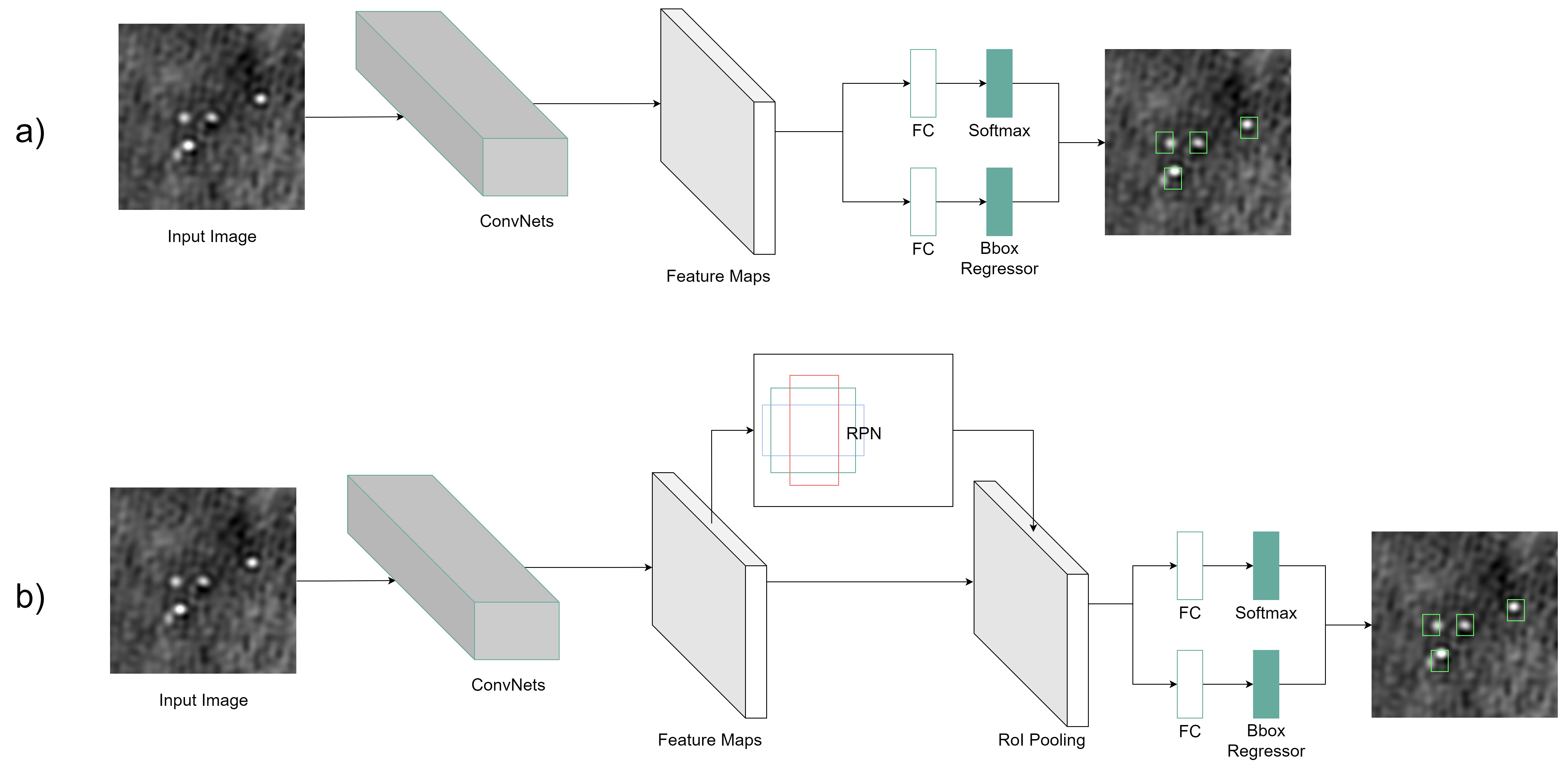}
	\centering
	\caption{General pipeline designs of a one-stage object detector a) and a two-stage object detector b), the latter employing RPN and RoI Pooling as additional operations to propose and filter regions.}
	\label{fig:od_arch}
\end{figure*}

\subsection{Detection Models}
\label{sec:detection_models}
Object Detection defines all deep neural networks designed with the goal of identifying and locating objects within an image. Such a task is carried out by defining bounding boxes and associating labels to the predicted object.
Deep learning-based approaches generally employ convolutional neural networks (CNNs) to perform end-to-end object detection~\cite{effdet,fasterrcnn,he2017mask,redmon2016you}, but, recently, transformer-based ones gained relevant popularity~\cite{detr,yolos,ddetr,tspdet}.
CNN-based object detection models typically consist of two modules: a backbone network and several prediction heads. The backbone generates a low-resolution, feature-rich image representation, usually employing widely affirmed classifier architectures (e.g. ResNets). Feature maps are then fed to a series of convolutional layers that learn to predict a bounding box and a label for each detected object.

Most object detection networks employ a region proposal network (RPN), introduced in~\cite{fasterrcnn}. This network generates a set of $N$ bounding boxes at different aspect ratios for each point on the convolutional feature map. The output of the backbone network is then used to determine whether each bounding box belongs to the foreground or the background by providing the feature map cropped by the anchor box to a small CNN. Then, starting from these anchor boxes, the objective function is computed on the offset between the anchor boxes and the ground truth. Usually, multiple proposals will yield a high score for the same object, especially for large objects, so further filtering is needed. Non-Maximum Suppression (NMS) computes the Intersection over Union (IoU) between the highest-scoring predicted box and the next high-scoring boxes and removes the ones with IoU higher than a threshold. Such a threshold is a hyperparameter. This family of models is known as two-stage detectors, as they process the input in two separate phases.

One-stage detectors skip the region proposal network by dividing the input image into $N_{g}$ grids. The model predicts $B$ bounding boxes and a confidence score for each grid. NMS is applied to the highest-scoring boxes to filter redundant predictions. These detectors allow for a more efficient prediction at inference time in terms of computational resource requirements, at the cost of model performance.

These methods limit the proposal of bounding boxes to a fixed number, which may seem a limitation. As we are working on image crops that contain fewer than fifteen objects, the number of proposed boxes is higher than the maximum number of objects.

Fig.~\ref{fig:od_arch} shows the general framework of these two families of object detectors.

Transformer-based methods employ a simpler approach by using the backbone network in the same way as CNN-based ones and feeding its output to a transformer~\cite{transformers}, as used in~\cite{detr}, or skip the use of the CNN backbone and use an encoder-only transformer~\cite{yolos}. The final prediction is given by a series of prediction heads, typically linear layers. The loss function employed in these models is a bipartite matching loss, introduced in~\cite{detr}. Transformers require more computational resources and are more difficult to lead to convergence, but can yield more robust models.
Note that some approaches have similar names and can generate confusion. YOLO (You Only Look Once) is a single-stage detector based on CNNs. YOLO9000, YOLOv3, and YOLOv4 are gradual improvements of the same architecture. YOLOS (You Only Look at One Sequence) is a separate model with a completely different architecture.

\subsubsection{Mask R-CNN}
\label{sec:asgard}
We used a Mask R-CNN-based model adapted for working on radio-astronomical images~\cite{maskrcnn-simone}. Mask R-CNN~\cite{he2017mask} is a deep learning model that builds on Faster R-CNN~\cite{fasterrcnn}, which, in turn, builds on R-CNN~\cite{girshick2014rich}.
Mask R-CNN performs instance segmentation on images, i.e. it combines object detection, classification, and semantic segmentation, in the form of a per-pixel mask for each object, differentiating between overlapping objects.

The first component of Mask R-CNN, which is also one of the newly introduced features compared to Faster R-CNN, is the Feature Pyramid Network (FPN)~\cite{fpn}. The FPN, also referred to as the backbone, serves as a feature extractor for objects at multiple scales, with shallower layers detecting lower-level features and deeper layers detecting higher-level features.
The bottom-up pathway is typically a ResNet~\cite{resnet} (ResNet101 in our case), chosen for their residual connections, which add output and input of a series of neural layers to solve the vanishing gradient problem~\cite{vanishing}. This problem relates to the gradients becoming too small in deeper layers due to chained multiplications by small numbers.
The top-down pathway consists of a convolution at each layer, followed by a lateral connection from the bottom-up pathway. This lateral connection allows the FPN to combine the top-down pathway's high-level features with the bottom-up pathway's low-level features.
The output of the FPN is then passed to the Region Proposal Network (RPN).

The RPN scans predefined areas of the image, referred to as anchors, and picks out regions likely to contain an object of interest. This component does not look directly at the input image again but uses features already extracted by the FPN, making it efficient. For each anchor, the RPN determines whether it is background or foreground, meaning whether it contains an object of interest and refinements, which are more precise `outlines' of the object. The regions proposed with the highest confidence are refined and passed on to the next component.

The regions proposed by the RPN are of varying shapes and sizes, which is problematic for the architecture of the network, and are thus resized to a fixed, predefined size. This step is called ROI Pooling. In Faster R-CNN, this is accomplished using ROIPool, which crops and resizes the feature map. However, this can result in the loss of significant data, along with spatial misalignment of mask pixels when overlayed with the original image. Mask R-CNN instead utilizes ROI Align, which samples the feature map at different points and uses bilinear interpolation to achieve the desired size.

The final component, and what particularly sets Mask R-CNN apart from its predecessors, is the Fully Convolutional Network (FCN), also known as the `mask branch'. This branch is a fully convolutional network that can retain spatial information and generates a per-pixel mask for the proposed regions and thus for the input image.

Given the nature of the data, we carry out some data preprocessing steps specific to radio astronomical data, which are not present in common implementations of Mask R-CNN. For example, setting `NaN' pixels to the image minimum and applying Z-Scale~\cite{zwitter2000introduction} normalization. Moreover, A significantly intensive Hyper-Parameter Optimization process is also carried out to fine-tune the model's architecture for the data it is made to handle. Furthermore, when being used for detection, we apply post-processing steps, such as merging overlapping or connected objects of the same label or retaining only the object with the highest confidence in other cases.

\subsubsection{Detectron2}
\label{sec:detectron}
Detectron2~\cite{detectron2} is Facebook AI Research’s next-generation library that implements
state-of-the-art object detection and instance segmentation algorithms. It is a rewrite of Detectron~\cite{detectron} that started with maskrcnn-benchmark~\cite{mrcnnbench}. It supports multiple tasks such as bounding box detection, instance segmentation, keypoint detection, densepose detection, and others. It also provides pre-trained models that can be easily loaded and used on new data sets. It allows for custom state-of-the-art computer vision technologies to be easily plugged in and includes robust models, e.g. Faster R-CNN~\cite{fasterrcnn}, Mask R-CNN~\cite{he2017mask}, RetinaNet~\cite{retinanet}, and DensePose~\cite{densepose}, and also features several new models, including Cascade R-CNN~\cite{cascadercnn}, Panoptic FPN~\cite{panopticfpn}, and TensorMask~\cite{tensormask}.

For detecting objects, our experiments made use of the Base (Faster) R-CNN with Feature Pyramid Network (Base-RCNN-FPN), which is the basic bounding box detector extendable to Mask R-CNN for instance segmentation. Faster R-CNN with FPN backbone is a multi-scale detector capable of detecting objects at different scales, accomplishing high accuracy and efficiency. A generic Region-Based Convolutional Neural Network (R-CNN) is composed of three main components. The first is a region proposal network that generates candidate regions (bounding boxes) using computer vision techniques. The second one is the feature extraction module which uses convolutional neural networks to extract the features from the candidate regions. Finally, the last component is a classifier that predicts the classes of the proposed candidates using the extracted features.

Specifically, for Faster R-CNN, these components are:
\begin{enumerate}
    \item Backbone Network: extracts feature maps from the input image at different scales. Base-RCNN-FPN output features are called P2 ($1 / 4$ scale), P3 ($1/8$), P4 ($1/16$), P5 ($1/32$) and P6 ($1/64$). 
    \item Region Proposal Network: detects object regions from the multi-scale features. $1000$ box proposals (by default) with confidence scores are obtained.
    \item Box Head: crops and warps feature maps using proposal boxes into multiple fixed-size features, and obtains fine-tuned box locations and classification results via fully-connected layers. Finally, $100$ boxes (by default) in maximum are filtered out using non-maximum suppression (NMS).
\end{enumerate}
    
Mask R-CNN extends on Faster R-CNN by adding another branch in parallel for pixel-level object instance segmentation. The branch is a fully connected network applied on ROIs to classify each pixel into segments with little overall computation cost. It adds a mask head parallel to the classification and bounding box regressor heads. With respect to Faster R-CNN, one major difference is the use of the RoIAlign layer, instead of the RoIPool layer, to avoid pixel-level misalignment due to spatial quantization. 

Mask R-CNN performs better than the existing state-of-the-art single-model architectures. It is simple to train, flexible, and generalizes well in many applications.

\subsubsection{YOLO}
\label{sec:yolo}
Characterized by a fully convolutional network that simultaneously predicts a set of bounding boxes and the associated class probabilities, YOLO \cite{redmon2016you} falls into the category of one-stage detectors. Such a model is characterized by faster execution time in comparison to two-stage detectors, at the cost of a certain degree of accuracy. The loss of accuracy is implicit in the design of the model, as YOLO detects objects by analyzing patches of the entire image. This causes a bias in the prediction of each object, which depends on the context of the patch where the object is located, meaning that the same object, put in another context, might be misdetected.
At the core of this method is the following idea: the input image, whose resolution is $P \times P$, is divided into an $S \times S$ grid, and each cell is treated as responsible for detecting an object if the center of such object falls in that cell. Note that $S << P$, so each grid cell contains multiple pixels. The authors set $S = 7$. Each cell predicts a fixed number $B$ of bounding boxes, and a confidence score for each of them, which determines how confident the box is to contain an object and also how accurate it is at covering the object. The confidence score is given by
\begin{equation}
    Pr \left( Object \right) \times IoU
\end{equation} 
where $Pr(Object)$ refers to the probability of that box containing an object, and $IoU$ measures the intersection over union between the prediction and the ground truth. This quantity should be as high as possible when an object is contained in the cell. If the model predicts no objects in a bounding box, the confidence score should be $0$.
$5$ values are predicted for each bounding box: $4$ coordinates ($xyhw$) to locate the bounding box in the image and the confidence score. Furthermore, for each grid cell, a class probability is computed for each category. Even if a grid cell contains multiple bounding boxes with different objects, the model will only predict a class for that cell, which translates into a loss of accuracy, especially for small objects in a group.

The architecture of YOLO is made up of $24$ convolutional layers and two fully connected layers. The first $20$ convolutional layers are pre-trained on the ImageNet \cite{russakovsky2015imagenet} dataset on the classification task; the whole pipeline is fine-tuned with the detection task (i.e. the predictions are bounding boxes). Training is carried out using the sum-squared error as the objective function, as it is easy to optimize.
Many grid cells do not contain any objects and will have a zero confidence score. This may cause an increase in the magnitude of the gradients from cells that do contain an object, leading to unstable training and divergence. To address this problem, the loss for the bounding box coordinate prediction is increased, while the loss relative to the confidence score is decreased for boxes that do not contain objects.
Although it represents an improvement in terms of performance and methodology when compared to two-stage approaches, YOLO presents the following limitations:
\begin{itemize}
  \item Limited flexibility since each grid cell predicts a fixed number of bounding boxes and a single class, which makes it impossible to distinguish among objects of different categories in the same grid cell or to predict more objects than the maximum number of boxes within a cell;
  \item Issues to generalize and recognize the same object in a different aspect ratio than the one provided in the training data, as its inductive bias is conditioned by the grid that divides the image;
  \item Errors are not treated proportionally to the box size but in terms of their absolute value, which is a problem as a small error on a large box does not affect prediction accuracy, as the same error value on a smaller box does.
\end{itemize}

There are multiple versions of this approach, each of which represents a relevant improvement in terms of computational performance and prediction accuracy relative to the older version. 
The first improvement of the architecture is represented by YOLO9000~\cite{yolov2}, which, among others, provides the following enhancements:
\begin{itemize}
    \item Batch normalization~\cite{batchnorm}, which contributes to stabilizing training;
    \item A higher resolution classifier, capable of recognizing features at a higher scale; 
    \item Multiscale training, which processes batches at different resolutions to make the model more robust to scale and aspect ratio variations. This is made possible by employing Feature Pyramid Networks (FPN)~\cite{fpn}.
\end{itemize} 
A further improvement in performance and computational speed is given by YOLOv3~\cite{yolov3}, which improves the backbone by adding residual connections and optimizes some training hyperparameters to make the model converge faster.
\changed{In our benchmark, we use two versions of this architecture: \textbf{YOLOv4} and the latest \textbf{YOLOv7}.} YOLOv4~\cite{yolov4} boosts YOLOv3 performance by operating on the following:
\begin{itemize}
    \item Expand the receptive field of the detector by increasing the number of convolutional layers in the backbone, integrating the Cross-Stage Partial Network~\cite{csp} into the backbone, and adding to this the Spatial Pyramid Pooling~\cite{spp};
    \item Replace the FPN in YOLOv3 with the PANet~\cite{panet} module 
\end{itemize}

\changed{YOLOv7~\cite{yolov7} builds upon this architecture and includes a series of architectural improvements to boost performance and resource requirements. Some of the improvements are the following:
\begin{itemize}
    \item Extended efficient layer aggregation to perform layer fusion that makes the operation less resource-demanding by reducing the path of the gradients in the computational graph;
    \item Auxiliary supervision heads in intermediate layers to improve the features' quality.
\end{itemize}
}

\subsubsection{EfficientDet}
\label{sec:effdet}
EfficientDet~\cite{effdet} represents a family of one-stage detectors whose general architecture follows that of YOLO and aims to improve efficiency and performance by exploring several architecture variations to reach state-of-the-art performance more efficiently.
This work realizes the enhancement by operating mainly on two aspects. First of all, most object detectors make use of multi-scale features to gather information from feature maps at different resolutions and fuse them to output the prediction. Such approaches do not distinguish between feature maps at different scales, so they treat them equally. Yet, each feature map highlights different properties of the image, so they should be processed accordingly. To tackle this problem, the authors introduce the Bi-directional Feature Pyramid Network (Bi-FPN). Scaling up the model performance is another concern addressed by this work. Most approaches rely on scaling up the backbone or the image size, while the authors claim it is necessary to scale the box and class prediction heads accordingly. Finally, they also modify the backbone architecture, choosing EfficientNet~\cite{effnet}, as it is less burdensome in terms of GFLOPs and model size.

The BiFPN module represents the core of the EfficientDet architecture and solves the task of gathering information from the image at different scales. This method improves existing state-of-the-art approaches like PANet~\cite{panet} and NAS-FPN~\cite{nas-fpn}, both requiring a high amount of resources to carry out training and inference, by employing a more efficient strategy to fuse features at different scales. In particular, they improve performance by operating on the following main aspects:
\begin{enumerate}
    \item Remove cross-scale connections that have only one input, as it has been observed that they contribute poorly to the final output by not providing a significant amount of information for the final prediction; 
    \item Add a residual connection between the input and the output for features at the same scale, so to fuse more features without adding parameters to the model;
    \item Repeat the top-down and bottom-up phases multiple times to gain a bidirectional flow of information, which enhances high-level feature fusion.
\end{enumerate}
    
Finally, as features need to be treated differently according to their resolution, the fusion strategy is modified. While most approaches use a rescaled sum of all features, this method offers the following ways to combine them:
\begin{itemize}
    \item Weighted fusion: feature maps are scaled by a learnable weight, and they are summed together. This may cause training instability, as the value of the weight is unbounded.
    \item Softmax-based fusion: each weight is given to a softmax operation, to rescale it into a range of $\left[0,1 \right]$. Computing softmax on each weight before multiplying it by the feature maps solves the aforementioned problem, but introduces a new one: significantly slower computation.
    \item Fast-normalized fusion: the operation of softmax is modified slightly by removing the exponential operation and adding an $\epsilon$ term to the denominator for numerical stability. This approach is more efficient and avoids the unbounded value problem.
\end{itemize}

The overall model consists of a backbone based on EfficientNet, a multi-scale feature aggregation network, carried out using BiFPN, and a box/class prediction network, realized by employing fully convolutional networks.

The model can be scaled up or down by setting a coefficient $\phi$, which influences the depth and width of the modules. For the backbone, the coefficient determines which EfficientNet architecture to use, from EfficientNet-B0 to B6. The BiFPN network is scaled according to the following equations:

\begin{equation}
    W_{BiFPN} = 64 \cdot \left (1.35^{\phi} \right ), \qquad
     D_{BiFPN} = 3 + \phi
\end{equation}

where $W_{BiFPN}$ indicates the number of channels of the convolutions (width) and $D_{BiFPN}$ refers to the number of layers in the network (depth).

The box and class predictors are scaled as well in the same manner. They share their width with the BiFPN module, but their depth is computed in the following way:

\begin{equation}
    D_{box} = D_{class} = 3 + \lfloor \phi / 3 \rfloor
\end{equation}

Finally, the input image resolution must also be scaled so that it can be processed by the BiFPN. For this reason, the input image size has to be divisible by $2^{7}$, thus the following equation:
\begin{equation}
    S_{input} = 512 + \phi \cdot 128
\end{equation}

The family of EfficientDet models includes EfficientDet-D0 to D7, where the number corresponds to the value of $\phi$.

We tested EfficientDetD1 and D2, as testing more complex versions of the models on our relatively simple images could lead to severe overfitting.

\subsubsection{DETR}
\label{sec:detr}
DETR~\cite{detr} is one of the first approaches to employ transformers to perform object detection. This method poses object detection as a direct set prediction problem, thus relieving the bounding box regression from the burden of Region Proposal \cite{fasterrcnn} and Non-Maximum Suppression. These two steps have a high impact on performance, as they consist in generating thousands of proposals and discarding most of them based on an Intersection over Union threshold value.
DETR consists of three main modules: (1) a CNN backbone, typically ResNet50 \cite{resnet}, which serves as a 2D feature extractor of the input image; (2) an encoder-decoder transformer, which computes self- and cross-attention on the features extracted by the backbone and produces a fixed set of predictions; (3) a feedforward network that assigns either the predicted class or a special \textbf{no-object} class to each element of the prediction set.

The core mechanism of the model resides in the transformer module, which is composed of an encoder and a decoder. Before being fed to the encoder, 2D features are projected onto a hidden dimension $d$ by means of a $ \times 1$ convolution and embedded with a spatial positional encoding. The transformer is designed to process sequences, so the 2D features of dimension $d \times H \times W$ are flattened into a sequence of $d\times HW$. Then, the encoder computes self-attention on the sequence followed by residual connections, layer normalization~\cite{layernorm}, and feedforward layers. Such operations are iterated as many times as the number of encoder layers stacked together. The decoder shares part of its architecture with the encoder but is characterized by an additional module: the cross-attention module, placed after the self-attention block. The decoder receives a fixed set of \emph{object queries} as input, which are randomly initialized learnable parameters. Then it computes self-attention on the input, followed by a cross-attention between the object queries and the encoder output. This step is crucial in transformer architectures as it contributes to mapping the information between the source sequence (i.e. the image features) and the target sequence (i.e. the predicted object categories and locations).

Finally, each element of the output sequence from the decoder is fed to a shared feedforward network to predict the bounding boxes and the corresponding class. The decoder always outputs a fixed set of predictions, equal to the number of provided object queries, and each prediction is given a certain class if an object is predicted with a high confidence score, otherwise, the object is associated with a \textbf{no-object} class.

An important remark about the decoder's behavior is that it differs from the one defined in~\cite{transformers}, which operates by iteratively generating an element of the sequence and feeding that element back to the decoder to generate the next element. In DETR, the decoder executes in a single step, predicting all the outputs in parallel.
Since the problem is framed as a direct set prediction one, the loss function to optimize is a matching cost between the ground truth set and the predicted set. The loss optimization follows two steps: (1) the Hungarian algorithm is used to find the optimal assignment among all the possible permutations of the two sets; (2) the Hungarian loss gives the actual loss function between the sets. The loss functions also depend on the bounding box loss, which uses a linear combination of the $L1$ loss and the IoU loss~\cite{iou_loss}. Additionally, an auxiliary loss is employed at each decoding layer to improve performance.

The main contribution of this work is that, by using attention maps, DETR can more accurately predict complex and overlapping objects, avoiding the computational burden that comes with traditional two-stage detectors.
A typical drawback of this model is its non-trivial training, which requires pre-training on extended datasets and learning rate warm-up schedules, as reported in \cite{difficulty_training_trf}, and its limited capability on smaller objects.

We tested the minimal version of DETR among the variants reported in \cite{detr}, i.e. DETR with ResNet50 as the backbone for feature extraction.

\subsubsection{YOLOS}
\label{sec:yolos}
Finally, to complete the overview of the existing models, we also tested a novel object detection approach based entirely on transformers. While DETR still relies on a convolutional backbone, YOLOS employs a fully transformer-based architecture similar to ViT~\cite{vit}, adapted to the object detection task, as the latter is designed to tackle classification tasks. Compared to ViT, YOLOS drops the \emph{[CLS]} token, as its purpose is to classify the input by performing attention on the whole patch sequence, whereas here, the task is to detect objects in the scene and locate them. Instead of using the \emph{CLS} token, YOLOS adds a hundred learnable detection tokens concatenated to the input image patches along the sequence dimension. The encoder then performs self-attention over the whole sequence, and the tokens will be used as output features that will be processed to generate the predictions. These tokens have the same purpose as the object queries in DETR.  
YOLOS shares some aspects with DETR: in addition to the concept of learnable tokens and the core based on transformers, they employ the same optimization strategy by minimizing a bipartite loss function, using the Hungarian algorithm and the respective loss defined in~\cite{detr}. Yet, YOLOS introduces a significant change in the transformer architecture: it employs an encoder-only transformer, while DETR presents an encoder-decoder architecture. The lack of a decoder module allows for a smaller number of parameters, thus a lighter model in inference, without significantly impairing prediction performance. 
Before being fed to the transformer, the image is divided into patches of fixed dimensions. Then each patch is flattened to have a sequence of one-dimensional vectors and projected onto a latent space. A fixed-size sequence of learnable detection tokens, each of which has the same size as the projected patches, is concatenated to the input sequence. This composed sequence is then summed with a sinusoidal positional encoding to maintain positional information within the sequence in the transformer.
The transformer computes self-attention between the elements of the input sequence, which in this case is the union of the projected patches and the detection tokens. This step is necessary for injecting information from the input image patches into the detection tokens.
The transformer encoder outputs a sequence of the same length as the input, but only the elements corresponding to the detection tokens are used to compute the predicted objects. Such output is provided as input to a 2-layer MLP with GELU~\cite{gelu} activation functions. Finally, two distinct heads, each consisting of a one-layer MLP with a ReLU activation function, predict the bounding box coordinates and categories, respectively.
A peculiarity of the approach used in YOLOS is the minimization of injected inductive bias. This means that the model exploits as little as possible the spatial structure and geometry of the input, which makes this approach versatile and capable of efficiently operating on data of different sizes and aspect ratios. The only form of spatial inductive bias implicit in the model is given by the patch sub-division of the image, but, other than this, no spatial convolutions are involved.

YOLOS comes in different variants, which differ mainly in model size, influenced by the embedding size and the number of heads. As data is not too complex, consisting of grayscale images, we tested the lightest version of the model, i.e. \textbf{YOLOS-Tiny}, to avoid having too many parameters, which might cause overfitting. This version consists of $12$ encoder layers, an embedding dimension of $192$, and $3$ heads for the attention mechanism, for a total of $5.7M$ parameters.

\subsection{Segmentation Models}
\label{sec:segmentation_models}
\begin{figure*}
	\centering
	\includegraphics[width=1\textwidth]{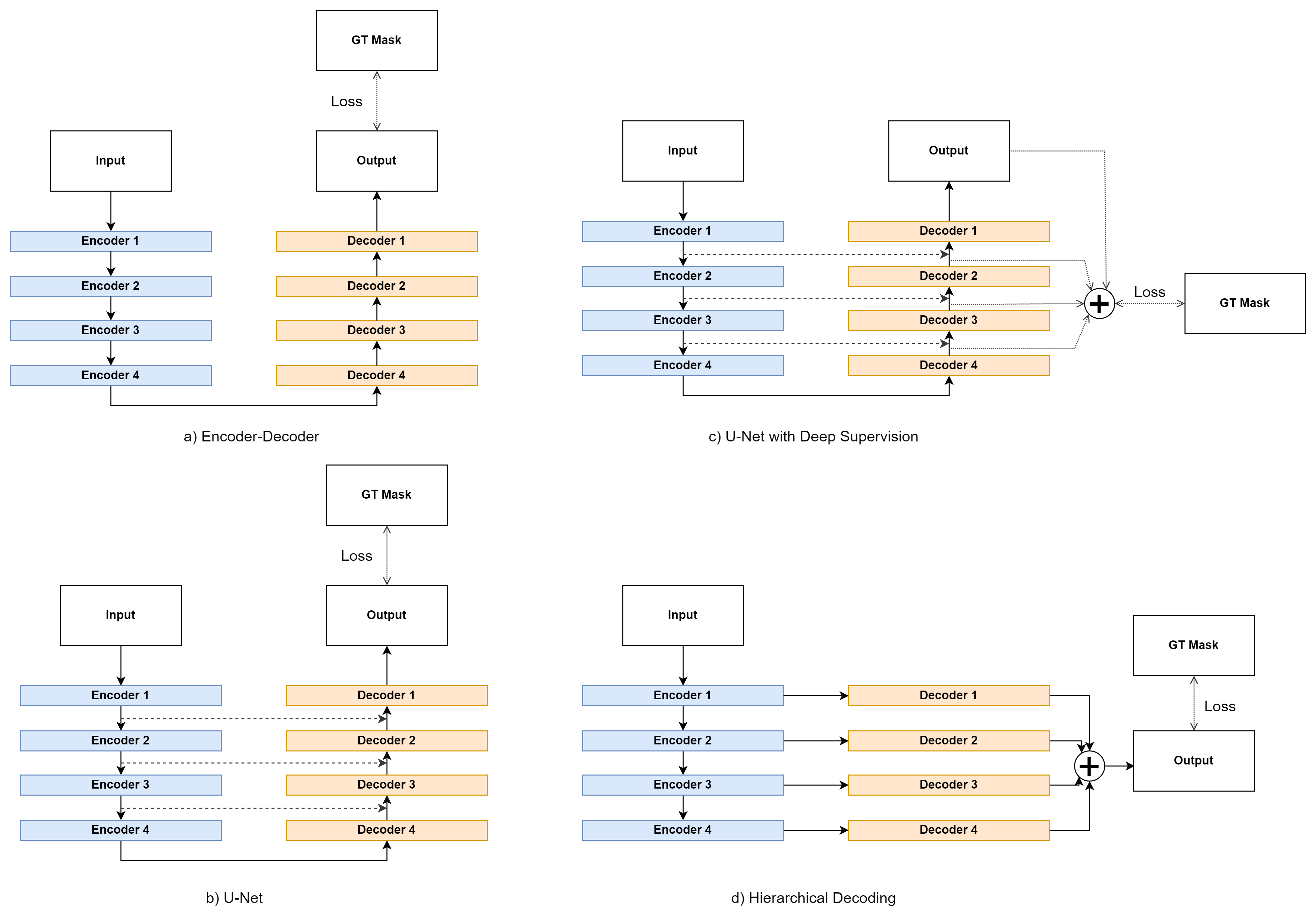}
	\caption{Different semantic segmentation architectures. a) Classic encoder-decoder approach; b) original U-Net architecture, with skip connections; c) U-Net with added deep supervision for enhanced loss; d) hierarchical decoding that combines features from all decoding stages.}
	\label{fig:seg_arch}
\end{figure*}

Semantic segmentation represents a task with the objective of classifying individual pixels of an image to obtain segmentation masks of the same size as the original image, where each pixel is labeled as belonging to its predicted class. This task is different from the object detection one, which predicts bounding boxes, so it generally follows a different pipeline. These models do not employ Region Proposal Networks since they do not need to ``find'' objects in the image, but classify each pixel. The objective function used to train these models is different as well; generally a form of the Cross-Entropy loss function, as opposed to L1 or MSE distances used in object detection.
The general pipeline used for semantic segmentation tasks consists of an encoder-decoder pair. The first represents the downsampling path, where the image is compressed to a latent space representation, while the latter is the upsampling path, which generates the high-resolution binary mask for the input image. In between the encoder and the decoder, a bottleneck layer projects the compressed image representation into the space of the binary mask by means of $1 \times 1$ convolutions.
In our analysis, we employ the following architectures on our radio-astronomical dataset, as shown in Fig.~\ref{fig:seg_arch}: a basic encoder-decoder architecture, U-Net~\cite{ronneberger2015unet}, which adds skip connections to the basic architecture, U-Net with deep supervision, which enhances the loss by computing it also on intermediate upsampled masks, Tiramisu~\cite{tiramisu}, which employs DenseNets~\cite{densenet} instead of convolutions in the downsampling and upsampling blocks, and PankNet~\cite{panknet}, which makes use of hierarchical decoding for information mixing between upsampling paths.

Segmentation models were originally designed for medical imaging applications. We think that our use case can benefit from such architectures, as medical images and radio-astronomical data are characterized by a similar distribution.

\subsubsection{Encoder-Decoder}
\label{sec:encdec}
The baseline for our segmentation models is a simple encoder-decoder pipeline with a bottleneck layer between the two paths, as shown in Fig.~\ref{fig:seg_arch}(a). This model employs a series of convolutional blocks in the encoder and several up-convolutional blocks in the decoder.
Each downsampling block is composed of two $3\times3$ convolutions followed by a rectified linear unit (ReLU) and a $2\times2$ max pooling to reduce feature resolution. The feature channels are doubled at each downsampling step. In the decoding path, each block performs a $2\times2$ transposed convolution, which expands the resolution of the feature map while halving the size of the feature channel, followed by $2$ layers of $3\times3$ convolutions and a ReLU. In the final layer, $1\times1$ convolutions are applied to map the dimension of the feature map to the number of classes.

\subsubsection{U-Net}
\label{sec:u_net}
Deep layers in neural networks are generally hard to train, yielding a model incapable of properly converging to an optimal point. The most common reason behind this behavior is the problem known as vanishing gradient~\cite{vanishing}, which can occur in deep pipelines and impairs the training procedure. ResNets~\cite{resnet} offer residual connections as a solution to this problem by summing the input and the output of a convolutional block. In this way, the gradients can flow through the shortcut path and avoid becoming too small.
U-Net~\cite{ronneberger2015unet} addresses the problem by applying skip connections, as shown in Fig.~\ref{fig:seg_arch}(b). Skip connections have the function of conditioning the upsampling output of the decoding blocks on the feature maps of the downsampling block of the same size. This is accomplished by concatenating the feature map produced by the downsampling block with the input to the corresponding upsampling block.

Another enhancement can be applied by intervening in how the cost of the objective function is computed. In the classic U-Net architecture, the loss is computed only on the last layer's output by applying a cross-entropy loss between this and the ground truth masks. We employ a U-Net variant in our segmentation models by applying the mechanism known as deep supervision~\cite{unetds}. It consists of computing the loss between the output of each upsampling path and the downsized ground truth mask to guide the decoder to generate meaningful masks from the early stages. An example of this architecture is shown in Fig.~\ref{fig:seg_arch}(c).

\changed{Finally, among the U-Net-based architectures, U-Net++~\cite{unetpp} improves the skip pathways and the deep supervision mechanism to improve performance and enhance gradient flow between the encoding and decoding paths. While U-Net simply concatenates the output of the downsampling layers with the corresponding upsampling ones, U-Net++ extends this operation by adding dense convolutional layers between the two paths, which yields a more meaningful encoding of the features as well as simplifying the computation of the gradients, thus stabilizing the training. U-Net++ employs deep supervision as well but, instead of applying it to intermediate outputs of the upsampling path, as done in~\cite{unetds}, it computes it on the outputs of the convolutional layers of the skip pathways. This enables model pruning at inference time, which makes this model adaptable to environments with constrained computational capabilities.}

\subsubsection{Tiramisu}
\label{sec:tiramisu}
While U-Net introduces skip connections to solve the vanishing gradient problem, Tiramisu~\cite{tiramisu} acts on the structure of the convolutional blocks to make feature maps more representational by injecting higher-resolution information into lower-resolution maps. This is achieved by employing DenseNets~\cite{densenet}, which replace the sum operation between the input and output of each convolutional block, typically used in ResNets, with concatenation. This contributes to reducing the network's parameters, as it reuses feature maps from earlier layers, freeing deeper layers from the need to learn redundant features.

\subsubsection{PankNet}
\label{sec:panknet}

Finally, we employ PankNet~\cite{panknet}, which does not make use of the downsampling-upsampling pipeline but applies encoders and decoders in a multilevel way, as shown in Fig~\ref{fig:seg_arch}(d). Each decoder receives only the corresponding encoder output as input, while the other versions concatenated it to the previous decoder output. In addition to this change in architecture, the output mask is obtained in a different way from what we have seen so far. Output feature maps from all the decoders are summed together, allowing for mixing global and local features, thus enabling a more context-aware prediction.
The model is designed to process 3D CT and MRI scans, but since our case study involves 2D images, we convert the 3D convolutions into 2D when using them in our dataset.

\section{Performance Analysis}

This Section presents the metrics and related results computed to evaluate the performance of each model on object detection (see Section  \ref{sec:detection_metrics}) and semantic segmentation (see Section \ref{sec:segmentation_metrics}). Additionally, in Section \ref{sec:snr}, we present an evaluation of the ability of the models to perform with different data samples according to their signal-to-noise ratio (SNR).
Finally, we evaluate the performance of the models from the computational point of view (see Section \ref{sec:computing_performance}).
We report the metrics on the entire dataset and the \emph{extended} category. We are more interested in how the models perform for this particular category as the traditional algorithms are well-affirmed for compact source detection but less efficient on single-island extended sources and incapable of detecting multi-island extended sources. Also, we expect common deep learning models to underperform on compact and spurious sources due to their small size and irregular shape. While analyzing the performance, it is useful to recall the number of objects per class (Table~\ref{tab:data_split}) and per SNR bin (Table~\ref{tab:snr}). For further comparison, we report metrics also for the \emph{compact} category.
\subsection{Object Detection}
\label{sec:detection_metrics}
\subsubsection{Detection Metrics}
We evaluate the performance of several detection models on the radio images dataset and compute the following metrics:
\begin{itemize}
    \item Reliability (Precision)
    \begin{equation}
        {Reliability} = \frac{{TP}}{ {TP} + {FP} }
        \label{eq:precision}
    \end{equation}
    \item Completeness (Recall)
    \begin{equation}
        {Completeness} = \frac{{TP}}{{TP} + {FN} }
        \label{eq:recall}
    \end{equation}
    \item F1-Score
    \begin{equation}
        {F1} = \frac{2 \times {R} \times {C}}{{R} + {C}} = \frac{2 \times {TP}}{\left(2 \times {TP} \right) + {FP} + {FN}}
        \label{eq:f1}
    \end{equation}
    where we refer to R for Reliability and C for Completeness \\
    \item mAP@50
    \begin{equation}
        {mAP} = \sum_{c}{{AP}_c}
        \label{eq:map}
    \end{equation}
\end{itemize}

To compute such metrics, we first need to compute the number of true positives \emph{(TP)}, false positives \emph{(FP)}, and false negatives \emph{(FN)}. In the object detection context, a true positive is the combination of a predicted bounding box having an Intersection over Union \emph{(IoU)} greater than a specified threshold and a correctly predicted category for the detected object. \emph{IoU} is defined as the ratio between the intersection of the predicted bounding box and the corresponding ground truth and the union of the two. If this ratio is greater than the threshold and the class associated with the bounding box is correct, the prediction is a true positive. If the \emph{IoU} is less than the threshold, or the object is misclassified, we count a false positive. If the ground truth bounding box has no associated prediction, we consider it a false negative. In our case, we use an \emph{IoU} threshold of $0.9$ for reliability, completeness, and f1-score metrics. For \emph{mAP@50}, we use a threshold of $0.5$, as specified in the name. We chose a different threshold for \emph{mAP@50}, as this is the most common metric used to evaluate object detection models.
The Mean Average Precision at $0.50$ \emph{mAP@0.5}, defined in Equation~\ref{eq:map}, indicates the mean over the average precision (or reliability) for all classes above the specified threshold of $0.5$. To compute such a metric, predictions need first to be ranked, in descending order, and by confidence score value. This is the reason why we can't use this kind of metric for segmentation models, as they intrinsically lack a confidence score for the predicted object. After ranking the $N$ predictions, starting from the one with the highest confidence score, we compute the area under the precision-recall curve for each class $c$, that we indicate with $AP_{c}$. Finally, we compute the mean of the average precision metrics over all classes. 

We trained our models with different batch sizes, as some models require higher memory consumption, especially the ones using transformers. The batch size determines how the model is updated: a batch size of $B$ means that the model parameters are updated after performing the forward pass on $B$ elements.
We trained our detection models for 300 epochs, using, for testing each model, the weights yielding the best validation loss.

\subsubsection{Detection Results}
In Table~\ref{tab:det_results}, we report the results for the described models. Looking at the reliability column, YOLOv4 appears to be the best-performing model, but this high score is counterbalanced by low completeness $(\sim 0.5)$. This means that, while the model yields predictions with an \emph{IoU} score of more than $0.9$, such precise predictions concern half of the true objects in the images. The model is missing half of the predictions, so the reliability metric is not enough to evaluate performance. This particular case (high reliability, low completeness) is also the effect of choosing a high \emph{IoU} threshold $(0.9)$ which makes all the predictions with an \emph{IoU} below that threshold be considered false negatives, thus increasing the denominator of the Completeness formula (Equation~\ref{eq:recall}). 

A more indicative metric that takes into account both reliability and completeness is the f1-score, so it seems more reasonable to use this metric as a comparison between the methods. 

All the methods have been trained from scratch, except DETR, Detectron2, and YOLOS. These models have been fine-tuned starting from weights pre-trained on the COCO \cite{coco} dataset. We refer to fine-tuning as training the whole model end-to-end, initialized with pre-trained weights instead of random weights. The choice of not training DETR and YOLOS from scratch stems from the difficulty of reaching convergence with transformers when trained from scratch, compared with the case where it is fine-tuned, as shown in Table~\ref{tab:pretraining}. Detectron2 has been pre-trained for comparison with Mask R-CNN, which, in contrast, is trained from scratch. This way, we evaluate the impact of training the model starting from learned features against training from randomly initialized weights.

\begin{table*}[]
    \caption{Detection metrics. YOLOv4 shows the best reliability, but this high value is given by a high IoU threshold. BS stands for batch size}
    \begin{adjustbox}{max width=\textwidth}
    \begin{tabular}{|l|l|ccc|ccc|ccc|c|}
    \hline
        \multicolumn{1}{|c|}{\multirow{2}{*}{Model}} & \multicolumn{1}{c|}{\multirow{2}{*}{BS}} & \multicolumn{3}{c|}{Reliability} & \multicolumn{3}{c|}{Completeness} & \multicolumn{3}{c|}{F1-Score} & \multicolumn{1}{l|}{mAP} \\
        \multicolumn{1}{|c|}{} & \multicolumn{1}{c|}{} & \multicolumn{1}{l}{Compact} & \multicolumn{1}{l}{Extended} & \multicolumn{1}{l|}{Total} & \multicolumn{1}{l}{Compact} & \multicolumn{1}{l}{Extended} & \multicolumn{1}{l|}{Total} & \multicolumn{1}{l}{Compact} & \multicolumn{1}{l}{Extended} & \multicolumn{1}{l|}{Total} & \multicolumn{1}{l|}{Total} \\ \hline
        Mask R-CNN & 32 & 48.7\% & 88.8\% & 52.0\% & 82.3\% & 77.0\% & 79.5\% & 61.2\% & 82.5\% & 62.9\% & 70.2\% \\
        Detectron2 & 64 & 59.8\% & 62.9\% & 59.1\% & \textbf{83.7\%} & \textbf{90.9\%} & \textbf{83.9\%} & 69.7\% & 74.3\% & 69.4\% & \textbf{83.9\%} \\
        DETR & 2 & 75.0\% & 84.6\% & 76.4\% & 76.6\% & 84.9\% & 76.8\% & \textbf{75.8\%} & 84.8\% & \textbf{76.6\%} & 79.0\% \\
        Yolo v4 & 64 & \textbf{97.4\%} & \textbf{95.9\%} & \textbf{97.2\%} & 48.3\% & 85.5\% & 50.2\% & 64.5\% & \textbf{90.4\%} & 66.2\% & 53.8\% \\
        \changed{Yolo v7} & \changed{32} & \changed{87.5\%} & \changed{87.7\%} & \changed{87.4\%} & \changed{60.0\%} & \changed{86.6\%} & \changed{61.0\%} & \changed{69.1\%} & \changed{87.2\%} & \changed{71.7\%} & \changed{61.6\%} \\
        YOLOS & 2 & 55.9\% & 78.1\% & 58.0\% & 75.0\% & 84.8\% & 75.5\% & 64.1\% & 81.3\% & 65.6\% & 76.3\% \\
        EffDet-D1 & 64 & 96.1\% & 0.0\% & 64.9\% & 42.2\% & 0.0\% & 33.7\% & 58.6\% & 0.0\% & 44.4\% & 53.5\% \\
        EffDet-D2 & 32 & 96.7\% & 0.0\% & 69.8\% & 48.5\% & 0.0\% & 39.1\% & 64.6\% & 0.0\% & 50.1\% & 53.8\% \\ \hline
    \end{tabular}
    \end{adjustbox}
    \label{tab:det_results}
\end{table*}

Detectron2 and DETR yield the best results, but this comes at a high computational cost (see Section~\ref{sec:computing_performance}). If a compromise between performance and computational budget has to be met, the single-stage detector YOLO or one of the EfficientDet versions may be more suitable.

\begin{table*}[]
    \centering
    \caption{Impact of pretraining for transformer-based models. FT stands for fine-tuning: "No" means the model has been trained from scratch, and "Yes" that it is fine-tuned from weights on the COCO Dataset.}
    \begin{tabular}{|lccccc|}
    \hline
        Model & FT & Rel & Comp & F1 Score & mAP@50 \\ \hline
        \multirow{2}{*}{DETR} & No  & 15.3\%  & 18.4\%  & 16.7\%  & 13.4\%  \\ 
        ~ & Yes  & \textbf{76.4\%}  & \textbf{76.8\%}  & \textbf{76.6\%}  & \textbf{78.9\%}   \\ \hline 
        \multirow{2}{*}{YOLOS} & No  & 48.1\%  & 25.9\%  & 33.7\%  & 26.4\%   \\ 
        ~& Yes  & \textbf{58.0\%}  & \textbf{75.5\%}  & \textbf{65.6\%}  & \textbf{76.3\%}   \\ 
    \hline
    \end{tabular}
    \label{tab:pretraining}
\end{table*}

The models perform a lower score in reliability and completeness \changed{compared to computer-vision-based techniques (e.g. \textsc{caesar}~\cite{riggi2019caesar}, AEGEAN~\cite{aegean})} but they come with the capability of distinguishing between multiple kinds of sources and detecting extended sources and artifacts. This is the main advantage of using deep learning for source detection, \changed{as shown by other object detection methods applied to radio astronomy, for instance, CLARAN~\cite{wu2019radio} and HeTu~\cite{hetu}}. Also, computer vision algorithms compute metrics regardless of the predicted object's category, and, in our case, this can be the reason why our performance is lower than \changed{this type of approach.}

In Section~\ref{sec:snr}, we also analyze the performance of these approaches on different Signal-to-Noise Ratio ranges to explore how much the models are capable of dealing with hard-to-detect sources and understand how much faint sources impair detection performance.

\subsubsection{\changed{Results by telescope type}}
\label{sec:subsets}
\changed{We perform an analysis of the impact of the image origin on the bias inducted in the models. We split our  dataset into three subsets, by telescope type: VLA, ATCA, and ASKAP, in the same way as in~\cite{maskrcnn-simone}. We then evaluate our best-performing object detector, YOLOv7, by creating three separate instances of the model, originating from different training sessions, one for each subset. We test each instance of the model on each subset and on the mixed dataset. We report the performance in terms of F1-score in Table~\ref{tab:subsets}
From this analysis, it emerges that each subset injects some degree of bias in the model, and this seems to be especially the case for the VLA subset. The model trained on single subsets yields poor performance when tested on different subsets, due to the distribution shift caused by the different telescopes of origin. When trained on the whole dataset, the model learns the biases of several telescopes, being capable of a higher degree of generalization}.

\begin{table*}[]
    \centering
    \caption{\changed{Performance on YOLOv7 when trained on different subsets of our dataset. Each row refers to an instance of YOLOv7 trained on the specified subset, while columns (2-5) specify the subset where each instance has been evaluated. "Mixed" stands for the whole dataset. Results are reported in terms of F1-score}}
    \begin{tabular}{|>{\color{black}}l|>{\color{black}}c>{\color{black}}c>{\color{black}}c>{\color{black}}c|}
    \hline
        \backslashbox{Train}{Test} & Mixed & VLA & ATCA & ASKAP \\ \hline
        Mixed & \textbf{71.7\%}  & \textbf{87.09\%}  & \textbf{90.25\%} & \textbf{87.04\%} \\ 
        VLA & 18.56\%  & 84.43\%  & 0.00\% & 0.27\%  \\ 
        ATCA & 38.83\%  & 01.32\%  & 59.85\% & 51.26\% \\ 
        ASKAP & 51.26\%  & 02.40\%  & 48.03\% & 81.33\% \\ 
    \hline
    \end{tabular}
    \label{tab:subsets}
\end{table*}

\subsection{Semantic Segmentation}
\label{sec:segmentation_metrics}
\subsubsection{Segmentation Metrics}
\begin{table*}[]
\caption{Semantic Segmentation metrics}
    \begin{adjustbox}{max width=\textwidth}
    \begin{tabular}{|ll|ccc|ccc|ccc|}
    \hline
        \multicolumn{2}{|c|}{\multirow{2}{*}{Model}} & \multicolumn{3}{c|}{Reliability} & \multicolumn{3}{c|}{Completeness} & \multicolumn{3}{c|}{F1-Score} \\
        \multicolumn{2}{|c|}{} & \multicolumn{1}{l}{Compact} & \multicolumn{1}{l}{Extended} & \multicolumn{1}{l|}{Total} & \multicolumn{1}{l}{Compact} & \multicolumn{1}{l}{Extended} & \multicolumn{1}{l|}{Total} & \multicolumn{1}{l}{Compact} & \multicolumn{1}{l}{Extended} & \multicolumn{1}{l|}{Total} \\ \hline
        \multicolumn{2}{|l|}{U-Net (no-skip)} & 72.2\% & 84.4\% & 76.8\% & 65.2\% & 70.5\% & 68.5\% & 68.5\% & 76.8\% & 72.4\% \\
        \multicolumn{2}{|l|}{U-Net} & 79.4\% & 86.7\% & 82.5\% & 70.4\% & 76.1\% & 72.5\% & 74.6\% & 81.1\% & 77.1\% \\
        \multicolumn{2}{|l|}{U-Net+DS} & 79.8\% & \textbf{93.8\%} & 83.1\% & 74.3\% & 88.3\% & 76.5\% & 77.0\% & 91.0\% & 79.7\% \\
        \multicolumn{2}{|l|}{\changed{U-Net++}} & \changed{81.4\%} & \changed{91.5\%} & \changed{84.7\%} & \changed{82.4}\% & \changed{90.1}\% & \changed{80.4}\% & \changed{79.5\%} & \changed{92.7\%} & \changed{84.3\%} \\
        \multicolumn{2}{|l|}{Tiramisu} & \textbf{82.2\%} & 90.9\% & \textbf{86.5\%} & \textbf{90.4\%} & \textbf{97.1\%} & \textbf{91.4\%} & \textbf{86.1\%} & \textbf{93.9\%} & \textbf{88.9\%} \\
        \multicolumn{2}{|l|}{PankNet} & 77.6\% & 85.4\% & 79.5\% & 68.1\% & 75.8\% & 72.5\% & 72.5\% & 80.3\% & 75.8\% \\ \hline
    \end{tabular}
    \end{adjustbox}
    \label{tab:seg_results}
\end{table*}

Similarly to how we computed the metrics for the detection models, we introduce reliability, completeness, and the f1 score for the segmentation models. These quantities are computed in the same way, as reported in Equations~\ref{eq:precision},~\ref{eq:recall},~\ref{eq:f1}. The main difference between the metrics for the two families of models resides in how we compute true positives, false positives, and false negatives. While for object detection we consider the \emph{IoU} between the predicted and ground truth boxes to determine the nature of the prediction, for segmentation models, we compare masks.
If the model predicts, for pixel $i$, a class of $\hat{k}$, $k$ is the true class, we will count a true positive if $k = \hat{k}$, a false positive if $k \neq \hat{k}$, and a false negative if $k \neq 0$ and $\hat{k} = 0$.
Given the different nature of the metrics, it is not informative to compare object detection models directly with semantic segmentation models, so we will evaluate them separately.
We trained our models with a batch size of $32$, using Adam~\cite{adam} as an optimizer, with a learning rate of $10^{-4}$, using a cross-entropy loss function. Training has been carried out for 300 epochs, and we selected the weights for the model at the minimum validation loss for evaluation.

\subsubsection{Segmentation Results}
Table~\ref{tab:seg_results} reports the comparison between our segmentation models.
These results show how the performance increases when progressively adding incremental enhancements, starting from the baseline encoder-decoder model up to the Tiramisu architecture, as we described in the previous sections. 
This demonstrates that on radio-astronomical data, it is possible to achieve the same improvements that would have been obtained on a canonical, more affirmed dataset. 
The lower performance achieved by the last model, PankNet, can be due to the following factors: 1) the change in the decoder architecture, replacing the upsampling path with hierarchical decoding; 2) the fact that the model is originally designed for segmenting 3D volumes instead of 2D images.
\changed{From this analysis it emerges how these models perform better on extended sources with respect to compact sources. This can be caused by the fact that, dealing with deep networks, these can lose more fine-grained details in the feature extraction process, resulting in smaller objects not being detected. The difficulty of detecting smaller objects, especially in the semantic segmentation task, is a common problem~\cite{small1,small2,small3} and this can be exacerbated by the fact that in our case we are using image crops at $132 \times 132$ resolution, causing smaller compact sources information to be encoded in a restricted number of pixels.}

\subsection{SNR Performance}
\label{sec:snr}
In addition to performance in the entire dataset, we evaluated the ability of the models to perform with specific data samples. We divided our test set into three subsets based on their signal-to-noise ratio (SNR), calculated by dividing the peak flux value of each cutout by the amount of background noise, as shown in Equation~\ref{eqn:snr}.
We selected $2$ thresholds for the SNR to split our test set: $5$ and $20$. The first one is a commonly used threshold in astrophysics to determine a lower bound for source filtering, i.e. sources below this SNR value are not considered. We set this first threshold to explore how models can predict sources below such an SNR value, thus on difficult samples. The threshold of $20$ is arbitrary, with the sole purpose of further exploring how much the SNR value affects the performance of the models.
Such a split generates a class imbalance, yet this does not affect training, as we train the models on the whole dataset and use this split only for evaluation, thus in inference.
In particular, in the first two splits (i.e. SNR~$<$~5 and 5~$<$~SNR~$<$~20), there are no spurious sources, so performance for this class in such subsets will necessarily be 0.
\begin{table}[!ht]
    \caption{Object category distribution among SNR splits. The first two splits lack a significant number of spurious sources, so no significant result should be expected for such splits}
    \centering
    \begin{tabular}{|c|c|c|c|}
    \hline
        Category & SNR $<$ 5 & 5 $<$ SNR $<$ 20  & SNR $>$ 20\\ \hline
        Spurious & 0 & 0 & 267 \\
        Compact & 175 & 1505 & 4206 \\ 
        Extended & 5 & 156 & 1046 \\ \hline
    \end{tabular}
    \label{tab:snr}
\end{table}

\begin{equation}
    SNR = \frac{F_P}{N_{BG}}
\label{eqn:snr}
\end{equation}

where $N_{BG}$ is the 3 Sigma Clip of the background pixels and $F_P$ is the peak flux computed on pixel values. The SNR equation that we used is defined in \cite{snr}.

\begin{table*}[!ht]
\caption{Performance evaluation on different SNR bins for each class. Results are expressed as f1 scores.}
\begin{adjustbox}{max width=\textwidth}
\begin{tabular}{|llccccccccc|}
\hline
\multicolumn{11}{|c|}{Detection Models} \\ \hline
\multicolumn{2}{|c|}{\multirow{2}{*}{Model}} & \multicolumn{3}{c|}{SNR \textless 5} & \multicolumn{3}{c|}{SNR 5 - 20} & \multicolumn{3}{c|}{SNR \textgreater 20} \\
\multicolumn{2}{|c|}{} & \multicolumn{1}{l}{Compact} & \multicolumn{1}{l}{Extended} & \multicolumn{1}{l|}{Total} & \multicolumn{1}{l}{Compact} & \multicolumn{1}{l}{Extended} & \multicolumn{1}{l|}{Total} & \multicolumn{1}{l}{Compact} & \multicolumn{1}{l}{Extended} & \multicolumn{1}{l|}{Total} \\ \hline
\multicolumn{2}{|l|}{Mask R-CNN} & 45.2\% & 66.7\% & \multicolumn{1}{c|}{45.0\%} & 62.0\% & 79.5\% & \multicolumn{1}{c|}{62.8\%} & 61.8\% & 83.1\% & 63.8\% \\
\multicolumn{2}{|l|}{Detectron2} & 70.8\% & 38.7\% & \multicolumn{1}{c|}{69.1\%} & 79.6\% & 65.5\% & \multicolumn{1}{c|}{77.7\%} & 66.3\% & 76.1\% & 66.9\% \\
\multicolumn{2}{|l|}{DETR} & 79.7\% & 66.7\% & \multicolumn{1}{c|}{79.0\%} & 83.0\% & 76.2\% & \multicolumn{1}{c|}{82.3\%} & \textbf{72.9\%} & 86.4\% & \textbf{75.4\%} \\
\multicolumn{2}{|l|}{YOLOv4} & 87.3\% & \textbf{80.0\%} & \multicolumn{1}{c|}{88.5\%} & 89.4\% & 91.6\% & \multicolumn{1}{c|}{90.6\%} & 72.0\% & 90.4\% & 74.6\% \\
\multicolumn{2}{|l|}{\changed{YOLOv7}} & \changed{\textbf{89.1\%}} & \changed{\textbf{80.0\%}} & \multicolumn{1}{c|}{\changed{\textbf{89.7\%}}} & \changed{\textbf{90.8\%}} & \changed{\textbf{92.1}\%} & \multicolumn{1}{c|}{\changed{\textbf{91.0\%}}} & \changed{72.4}\% & \changed{\textbf{91.7\%}} & \changed{75.2\%} \\
\multicolumn{2}{|l|}{YOLOS} & 51.3\% & 44.4\% & \multicolumn{1}{c|}{50.9\%} & 54.0\% & 49.7\% & \multicolumn{1}{c|}{50.9\%} & 41.7\% & 84.4\% & 43.6\% \\
\multicolumn{2}{|l|}{EffDet-D1} & 74.7\% & 0.0\% & \multicolumn{1}{c|}{73.2\%} & 70.6\% & 0.0\% & \multicolumn{1}{c|}{62.4\%} & 53.1\% & 0.0\% & 39.0\% \\
\multicolumn{2}{|l|}{EffDet-D2} & 76.2\% & 0.0\% & \multicolumn{1}{c|}{74.6\%} & 79.3\% & 0.0\% & \multicolumn{1}{c|}{71.7\%} & 57.9\% & 0.0\% & 43.3\% \\ \hline
\multicolumn{11}{|c|}{Segmentation Models} \\ \hline
\multicolumn{2}{|c|}{\multirow{2}{*}{Model}} & \multicolumn{3}{c|}{SNR \textless 5} & \multicolumn{3}{c|}{SNR 5 - 20} & \multicolumn{3}{c|}{SNR \textgreater 20} \\ 
\multicolumn{2}{|c|}{} & \multicolumn{1}{l}{Compact} & \multicolumn{1}{l}{Extended} &  \multicolumn{1}{l|}{Total} & \multicolumn{1}{l}{Compact} & \multicolumn{1}{l}{Extended} & \multicolumn{1}{l|}{Total} & \multicolumn{1}{l}{Compact} & \multicolumn{1}{l}{Extended} & \multicolumn{1}{l|}{Total} \\ \hline
\multicolumn{2}{|l|}{U-Net (no-skip)} & 82.1\% & \multicolumn{1}{r}{\textbf{66.7\%}} & \multicolumn{1}{c|}{80.5\%} & 68.0\% & 64.9\% & \multicolumn{1}{c|}{66.5\%} & 61.5\% & 84.8\% & 63.6\% \\
\multicolumn{2}{|l|}{U-Net} & \textbf{86.8\%} & \multicolumn{1}{r}{\textbf{66.7\%}} & \multicolumn{1}{c|}{\textbf{85.1\%}} & 73.3\% & 68.3\% & \multicolumn{1}{c|}{72.5\%} & 67.1\% & 87.0\% & 70.5\% \\
\multicolumn{2}{|l|}{U-Net+DS} & 83.9\% & \multicolumn{1}{r}{61.5\%} & \multicolumn{1}{c|}{82.5\%} & 73.8\% & 72.9\% & \multicolumn{1}{c|}{71.6\%} & 70.4\% & 89.6\% & 73.8\% \\
\multicolumn{2}{|l|}{\changed{U-Net++}} & \changed{84.4\%} & \multicolumn{1}{r}{\changed{\textbf{66.7\%}}} & \multicolumn{1}{c|}{\changed{84.8\%}} & \changed{78.6\%} & \changed{77.4\%} & \multicolumn{1}{c|}{\changed{75.9\%}} & \changed{72.3\%} & \changed{90.2\%} & \changed{75.2\%} \\
\multicolumn{2}{|l|}{Tiramisu} & 82.3\% & \multicolumn{1}{r}{\textbf{66.7\%}} & \multicolumn{1}{c|}{81.7\%} & \textbf{86.5\%} & \textbf{88.9\%} & \multicolumn{1}{c|}{\textbf{87.9\%}} & \textbf{90.7\%} & \textbf{91.6\%} & \textbf{91.9\%} \\
\multicolumn{2}{|l|}{PankNet} & 64.0\% & \multicolumn{1}{r}{63.16\%} & \multicolumn{1}{c|}{67.2\%} & 63.0\% & 74.0\% & \multicolumn{1}{c|}{67.9\%} & 70.0\% & 77.0\% & 75.8\% \\ \hline
\end{tabular}
\end{adjustbox}
\label{tab:snr_results}
\end{table*}

Table~\ref{tab:snr_results} reports the results computed on the different SNR bins. Models perform poorly in the range lower than $5$ because samples in this SNR range are more difficult to detect, even using visual inspection, and because, as shown in Table~\ref{tab:snr}, this bin contains a low number of samples. \changed{YOLOv7} seems to excel in this case among object detection models, while, in segmentation, Tiramisu and U-Net perform best for SNR values above and below $5$, respectively. \changed{YOLOv7} outperforms the other models in this scenario because of its high reliability at the $0.9$ IoU threshold, reported in Table~\ref{tab:det_results}, which raises the f1-score. Such results are enhanced by the prevalence of point sources rather than extended sources, suggesting a better ability of \changed{YOLOv7} to detect small objects compared to the other methods.
Among the segmentation methods, Tiramisu seems to be the most robust model, even if it may slightly overfit on fainter objects, for which U-Net seems better suited.

\subsection{Computational Resources Utilization}
\label{sec:computing_performance}
In this section, we evaluate the performance of the presented models from the computational point of view. We performed our experiments on a DGX cluster composed of $3$ nodes, each of which is equipped with $8$ \emph{NVIDIA A100} GPUs. We trained each model across 8 GPUs, using data parallelization by broadcasting a copy of the model and its parameters to each process, loading a shard of the training batches. At each forward pass, the gradients are reduced among all processes and the same backward pass is performed on all the copies. Such communication operations introduce some overhead that may limit the speed-up of the training process.

\begin{table}[!ht]
    \centering
    \caption{Computing performance on CINECA DGX machine. *ETA refers to one epoch}
    \begin{tabular}{|lccc|}
    \hline
        \multicolumn{4}{|c|}{\textbf{Detection models}} \\ \hline
        \multicolumn{1}{|c}{Model}  & ETA* & GFLOPs  & \# of params \\ \hline
        Mask R-CNN & 180s & 198.0 & 44M \\ 
        Detectron2 & 150s & 198.0 & 43.9M \\ 
        DETR & 130s & 85.5 & 41.5M \\ 
        YOLOv4 & 90s & 48.3 & 52.5M \\ 
        \changed{YOLOv7} & \changed{43s} & \changed{43.7} & \changed{29.3M} \\ 
        YOLOS & 75s & 22.7 & 5.7M \\ 
        EffDet-D1 & 52s & 6.1 & 6.6M \\ 
        EffDet-D2  & 77s & 11.3  & 8M \\ \hline
        \multicolumn{4}{|c|}{\textbf{Segmentation models}} \\ \hline
        \multicolumn{1}{|c}{Model}  & ETA* & GFLOPs  & \# of params \\ \hline
        Encoder-Decoder  & 50s & 14.9  & 7.0M \\ 
        U-Net & 56s & 16.6  & 7,7M \\ 
        U-Net + DS  & 67s & 19.8  & 7.7M \\ 
        \changed{U-Net++}  & \changed{75s} & \changed{27.4}  & \changed{9.0M} \\ 
        Tiramisu  & 97s & 32.4  & 3.5M \\ 
        PankNet  & 101s & 12.8  & 25.6M \\ \hline
    \end{tabular}
    \label{tab:comp_perf}
\end{table}

In Table~\ref{tab:comp_perf}, we show the computing performance on the CINECA DGX cluster in terms of the average time it takes to complete a training epoch. Along with this metric, we report GFLOPs (i.e. the number of billions of floating point operations that each model requires to make a forward pass) and the total number of parameters of each model.

Most current state-of-the-art models for object detection tasks employ a two-stage detection process based on the Region Proposal Network \cite{fasterrcnn}, which performs highly computationally demanding operations.
The high number of FLOPs required by Mask R-CNN and Detectron2 is due to their architecture, as they are based on the same model, i.e. Mask R-CNN \cite{he2017mask}. YOLOv4, being a single-stage detector, is more computationally efficient in comparison to Mask R-CNN-based ones, even if it requires more parameters.

DETR \cite{detr} comes with a less demanding set of operations compared to two-stage detectors (i.e. Mask R-CNN and Detectron2), as it doesn't use an RPN but still represents a burden on some architectures, as it employs transformers at its core. In particular, the use of both transformer encoder and decoder modules, with both self- and cross-attention mechanisms, is where most of the computational time is spent.

YOLOS uses a transformer architecture as well but requires less computational resources compared to DETR, mainly due to its encoder-only design, exploiting the sequence processing capability of the self-attention mechanism and avoiding the computational burden of the decoder cross-attention.

Among the segmentation approaches, all the U-Net-based ones and the baseline encoder-decoder model require approximately the same amount of parameters. Tiramisu is an exception, and, thanks to its feature reuse, given by its DenseNet blocks, allows for fewer parameters to be optimized in the training process.

\section{Conclusions}
\label{sec:conclusions}
With the advent of the Square Kilometre Array, automated techniques are needed to efficiently analyze the data deluge expected to be produced by this new generation of astronomical facilities. Deep learning is being evaluated in radio astronomy and shows promising results in source detection and classification. 

This work \changed{proposes a benchmark reporting performance and computational requirements of several models, both convolutional and transformer-based, for the tasks of object detection and semantic segmentation on radio astronomical images collected from different surveys and telescopes}. The results can guide future activities on several radio surveys based on the peculiarities of the available images and computing capacities. However, since deep learning is rapidly evolving, any newly developed model has been possibly omitted but additional releases of the present work are planned in the future, as well as extending the training dataset and classification capabilities. 
\changed{From the analyses we conducted, we noticed that for the object detection task, YOLO-based methods report better overall performance in comparison to other approaches. Transformer-based approaches, even if they show promising results, especially in the case of DETR~\cite{detr}, come with the drawback of computational burden, as the multi-head attention mechanism increases the number of operations. In addition to this, as shown in Table~\ref{tab:pretraining}, transformer models require a lot of data to be properly trained, e.g. 300M images~\cite{vit} which can be a problem when training in the radio-astronomical field, for the limited amount of labelled data available for training deep learning models. Also, for images in an SNR range lower than $20$, architectures based on YOLO showed the best performance, which can be ideal for cases where most of the images are characterized by lower peak fluxes. The comparison of semantic segmentation models shows that Tiramisu achieves the highest performance without excessively compromising computational speed. The major drawback of this kind of model lies in the format of the annotations required for training. Providing a segmentation mask that classifies each pixel for each image is not a trivial task, as these masks have to be carefully segmented by experts in the field, while bounding box coordinates are easier to provide.}

We also aim to address, in future works, the use of Generative Adversarial Network (GAN) architectures to solve a series of open problems in the radio astronomical field, such as the inability to replicate machine learning experiments due to data privacy and the difficulty of simulating annotated data in a non-time-consuming way.

\section*{Acknowledgements} 
The research leading to these results has received funding from the Horizon 2020 research and innovation programme of the European Commission under grant agreement No. 863448 (NEANIAS), from the INAF PRIN TEC programme (CIRASA), and from the MOSAICo (Metodologie Open Source per l’Automazione Industriale e delle procedure di CalcOlo in astrofisica) project, co-funded by the EU, Fondo Europeo di Sviluppo Regionale - Programma Operativo Nazionale Imprese e Competitività 2014-2020. \\
We thank the authors of the following software tools and libraries that have been extensively used in this work: \caesar{}~\citep{riggi2016caesar,riggi2019caesar}, astropy~\citep{astropy2013,astropy2018}, ds9~\citep{ds9}, \textsc{dvc}.
We also acknowledge the contribution made by Andrea Pilzer in improving this work.
This preprint has not undergone peer review or any post-submission improvements or corrections. The Version of Record of this article is published in Experimental Astronomy, and is available online at https://doi.org/10.1007/s10686-023-09893-w

\begin{appendices}

\section{Additional results}\label{sec:add_res}

\begin{table*}[ht]
\centering
    \caption{Reliability over SNR splits and over the compact and extended classes, as well as over all classes.}
    \begin{adjustbox}{max width=\textwidth}
    \begin{tabular}{|llccc|ccc|ccc|}
    \hline
        \multicolumn{11}{|c|}{Detection Models} \\ \hline
        \multicolumn{2}{|c|}{\multirow{2}{*}{Model}} & \multicolumn{3}{c|}{SNR \textless 5} & \multicolumn{3}{c|}{SNR 5 - 20} & \multicolumn{3}{c|}{SNR \textgreater 20} \\
        \multicolumn{2}{|c|}{} & \multicolumn{1}{l}{Compact} & \multicolumn{1}{l}{Extended} & \multicolumn{1}{l|}{Total} & \multicolumn{1}{l}{Compact} & \multicolumn{1}{l}{Extended} & \multicolumn{1}{l|}{Total} & \multicolumn{1}{l}{Compact} & \multicolumn{1}{l}{Extended} & \multicolumn{1}{l|}{Total} \\ \hline
        \multicolumn{2}{|l|}{Mask R-CNN} & 31.13\% & \multicolumn{1}{c}{66.67\%} & \multicolumn{1}{c|}{31.45\%} & 47.10\% & 74.59\% & \multicolumn{1}{c|}{51.15\%} & 50.10\% & 81.28\% & 53.18\% \\
        \multicolumn{2}{|l|}{Detectron2} & 57.40\% & 27.27\% & \multicolumn{1}{c|}{56.72\%} & 69.67\% & 58.46\% & \multicolumn{1}{c|}{68.81\%} & 53.73\% & 67.07\% & 57.51\% \\
        \multicolumn{2}{|l|}{DETR} & 73.30\% & 66.67\% & \multicolumn{1}{c|}{72.63\%} & 80.60\% & 72.78\% & \multicolumn{1}{c|}{79.79\%} & 72.89\% & 86.77\% & 75.32\% \\
        \multicolumn{2}{|l|}{YOLOv4} & \textbf{87.57\%} & \textbf{100.00\%} & \multicolumn{1}{c|}{\textbf{90.28\%}} & 95.03\% & 93.74\% & \multicolumn{1}{c|}{\textbf{94.76\%}} & 92.07\% & 94.16\% & 90.97\% \\
        \multicolumn{2}{|l|}{\changed{YOLOv7}} & \changed{86.91\%} & \textbf{\changed{100.00\%}} & \multicolumn{1}{c|}{\changed{89.16\%}} & \changed{92.52\%} & \changed{\textbf{93.91\%}} & \multicolumn{1}{c|}{\changed{93.64\%}} & \changed{94.54\%} & \changed{\textbf{94.35\%}} & \changed{\textbf{94.33\%}} \\
        \multicolumn{2}{|l|}{YOLOS} & 36.79\% & 50.00\% & \multicolumn{1}{c|}{36.88\%} & 38.83\% & 72.54\% & \multicolumn{1}{c|}{37.15\%} & 29.19\% & 84.20\% & 41.82\% \\
        \multicolumn{2}{|l|}{EffDet D1} & 86.59\% & 0.00\% & \multicolumn{1}{c|}{84.39\%} & 96.82\% & 0.00\% & \multicolumn{1}{c|}{82.11\%} & 96.38\% & 0.00\% & 58.45\% \\
        \multicolumn{2}{|l|}{EffDet D2} & 87.13\% & 0.00\% & \multicolumn{1}{c|}{85.44\%} & \textbf{97.72\%} & 0.00\% & \multicolumn{1}{c|}{86.54\%} & \textbf{96.71\%} & 0.00\% & 62.52\% \\ \hline
        \multicolumn{11}{|c|}{Segmentation Models} \\ \hline
        \multicolumn{2}{|c|}{\multirow{2}{*}{Model}} & \multicolumn{3}{c|}{SNR \textless 5} & \multicolumn{3}{c|}{SNR 5 - 20} & \multicolumn{3}{c|}{SNR \textgreater 20} \\
        \multicolumn{2}{|c|}{} & \multicolumn{1}{l}{Compact} & \multicolumn{1}{l}{Extended} & \multicolumn{1}{l|}{Total} & \multicolumn{1}{l}{Compact} & \multicolumn{1}{l}{Extended} & \multicolumn{1}{l|}{Total} & \multicolumn{1}{l}{Compact} & \multicolumn{1}{l}{Extended} & \multicolumn{1}{l|}{Total} \\ \hline
        \multicolumn{2}{|l|}{U-Net (no-skip)} & 86.16\% & 27.27\% & \multicolumn{1}{c|}{84.83\%} & 56.83\% & 76.85\% & \multicolumn{1}{c|}{54.89\%} & 54.37\% & 87.62\% & 64.81\% \\
        \multicolumn{2}{|l|}{U-Net} & 91.44\% & \textbf{66.67\%} & \multicolumn{1}{c|}{89.07\%} & 63.97\% & 78.54\% & \multicolumn{1}{c|}{62.17\%} & 62.51\% & 89.52\% & 67.91\% \\
        \multicolumn{2}{|l|}{U-Net+DS} & 91.59\% & \textbf{66.67\%} & \multicolumn{1}{c|}{\textbf{91.40\%}} & 63.52\% & \textbf{90.52\%} & \multicolumn{1}{c|}{65.13\%} & 68.44\% & 93.69\% & 73.48\% \\
        \multicolumn{2}{|l|}{\changed{U-Net++}} & \changed{\textbf{91.68\%}} & \changed{\textbf{66.67\%}} & \multicolumn{1}{c|}{\changed{\textbf{91.67\%}}} & \changed{71.38\%} & \changed{89.15\%} & \multicolumn{1}{c|}{\changed{68.73\%}} & \changed{71.39\%} & \changed{94.61}\% & \changed{76.31\%} \\
        \multicolumn{2}{|l|}{Tiramisu} & 83.18\% & 50.00\% & \multicolumn{1}{c|}{83.24\%} & \textbf{82.49\%} & 89.29\% & \multicolumn{1}{c|}{\textbf{84.64\%}} & \textbf{94.88\%} & \textbf{95.71\%} & \textbf{93.84\%} \\
        \multicolumn{2}{|l|}{PankNet} & 56.66\% & \textbf{66.67\%} & \multicolumn{1}{c|}{62.39\%} & 54.06\% & 82.17\% & \multicolumn{1}{c|}{57.13\%} & 66.81\% & 77.49\% & 75.36\% \\ \hline
    \end{tabular}
    \end{adjustbox}
    \label{tab:snr_rel}
\end{table*}

\begin{table*}[ht]
\centering
    \caption{Completeness over SNR splits and over the compact and extended classes, as well as over all classes.}
    \begin{adjustbox}{max width=\textwidth}
    \begin{tabular}{|llccc|ccc|ccc|}
    \hline
        \multicolumn{11}{|c|}{Detection Models} \\ \hline
        \multicolumn{2}{|c|}{\multirow{2}{*}{Model}} & \multicolumn{3}{c|}{SNR \textless 5} & \multicolumn{3}{c|}{SNR 5 - 20} & \multicolumn{3}{c|}{SNR \textgreater 20} \\
        \multicolumn{2}{|c|}{} & \multicolumn{1}{l}{Compact} & \multicolumn{1}{l}{Extended} & \multicolumn{1}{l|}{Total} & \multicolumn{1}{l}{Compact} & \multicolumn{1}{l}{Extended} & \multicolumn{1}{l|}{Total} & \multicolumn{1}{l}{Compact} & \multicolumn{1}{l}{Extended} & \multicolumn{1}{l|}{Total} \\ \hline
        \multicolumn{2}{|l|}{Mask R-CNN} & 82.52\% & \textbf{66.67\%} & \multicolumn{1}{r|}{78.95\%} & 90.82\% & 85.03\% & \multicolumn{1}{r|}{81.46\%} & 80.52\% & 84.98\% & 79.63\% \\
        \multicolumn{2}{|l|}{Detectron2} & \textbf{92.52\%} & \textbf{66.67\%} & \multicolumn{1}{r|}{88.47\%} & \textbf{92.82\%} & 74.49\% & \multicolumn{1}{r|}{\textbf{89.28\%}} & \textbf{86.69\%} & 87.85\% & \textbf{80.07\%} \\
        \multicolumn{2}{|l|}{DETR} & 87.28\% & \textbf{66.67\%} & \multicolumn{1}{r|}{86.52\%} & 85.53\% & 79.87\% & \multicolumn{1}{r|}{84.95\%} & 72.89\% & 85.95\% & 75.48\% \\
        \multicolumn{2}{|l|}{YOLOv4} & 87.06\% & \textbf{66.67\%} & \multicolumn{1}{r|}{86.71\%} & 84.46\% & \textbf{90.55\%} & \multicolumn{1}{r|}{86.72\%} & 59.15\% & 86.90\% & 63.19\% \\
        \multicolumn{2}{|l|}{\changed{YOLOv7}} & \changed{91.40\%} & \changed{\textbf{66.67\%}} & \multicolumn{1}{r|}{\changed{\textbf{90.25\%}}} & \changed{89.14\%} & \changed{90.36\%} & \multicolumn{1}{r|}{\changed{88.50\%}} & \changed{58.66\%} & \changed{\textbf{89.19\%}} & \changed{62.52\%} \\
        \multicolumn{2}{|l|}{YOLOS} & 84.95\% & 40.00\% & \multicolumn{1}{r|}{82.17\%} & 88.50\% & 37.81\% & \multicolumn{1}{r|}{80.61\%} & 72.86\% & 84.52\% & 45.56\% \\
        \multicolumn{2}{|l|}{EffDet D1} & 65.74\% & 0.00\% & \multicolumn{1}{r|}{64.55\%} & 55.51\% & 0.00\% & \multicolumn{1}{r|}{50.31\%} & 36.66\% & 0.00\% & 29.22\% \\
        \multicolumn{2}{|l|}{EffDet D2} & 67.69\% & 0.00\% & \multicolumn{1}{r|}{66.17\%} & 66.72\% & 0.00\% & \multicolumn{1}{r|}{61.26\%} & 41.33\% & 0.00\% & 33.14\% \\ \hline
        \multicolumn{11}{|c|}{Segmentation Models} \\ \hline
        \multicolumn{2}{|c|}{\multirow{2}{*}{Model}} & \multicolumn{3}{c|}{SNR \textless 5} & \multicolumn{3}{c|}{SNR 5 - 20} & \multicolumn{3}{c|}{SNR \textgreater 20} \\
        \multicolumn{2}{|c|}{} & \multicolumn{1}{l}{Compact} & \multicolumn{1}{l}{Extended} & \multicolumn{1}{l|}{Total} & \multicolumn{1}{l}{Compact} & \multicolumn{1}{l}{Extended} & \multicolumn{1}{l|}{Total} & \multicolumn{1}{l}{Compact} & \multicolumn{1}{l}{Extended} & \multicolumn{1}{l|}{Total} \\ \hline
        \multicolumn{2}{|l|}{Encoder-Decoder} & 78.46\% & 66.67\% & \multicolumn{1}{r|}{76.52\%} & 84.69\% & 56.18\% & \multicolumn{1}{r|}{84.18\%} & 70.88\% & 82.16\% & 62.40\% \\
        \multicolumn{2}{|l|}{U-Net} & \textbf{82.52\%} & 66.67\% & \multicolumn{1}{r|}{\textbf{81.50\%}} & 85.85\% & 60.36\% & \multicolumn{1}{r|}{86.89\%} & 72.52\% & 84.56\% & 73.27\% \\
        \multicolumn{2}{|l|}{U-Net+DS} & 77.45\% & 66.67\% & \multicolumn{1}{r|}{75.16\%} & 88.16\% & 61.08\% & \multicolumn{1}{r|}{79.42\%} & 72.45\% & 85.89\% & 74.12\% \\
        \multicolumn{2}{|l|}{\changed{U-Net++}} & \changed{78.19}\% & \changed{66.67\%} & \multicolumn{1}{r|}{\changed{78.89\%}} & \changed{87.44\%} & \changed{68.39\%} & \multicolumn{1}{r|}{\changed{84.74\%}} & \changed{73.23\%} & \changed{86.18\%} & \changed{74.12\%} \\
        \multicolumn{2}{|l|}{Tiramisu} & 81.52\% & \textbf{80.00\%} & \multicolumn{1}{r|}{80.22\%} & \textbf{90.81\%} & \textbf{88.57\%} & \multicolumn{1}{r|}{\textbf{91.44\%}} & \textbf{86.82\%} & \textbf{87.87\%} & \textbf{89.98\%} \\
        \multicolumn{2}{|l|}{PankNet} & 73.52\% & 66.67\% & \multicolumn{1}{r|}{72.88\%} & 75.48\% & 67.31\% & \multicolumn{1}{r|}{83.70\%} & 73.52\% & 76.51\% & 76.27\% \\ \hline
    \end{tabular}
    \end{adjustbox}
    \label{tab:snr_comp}
\end{table*}

We report additional results, in Table~\ref{tab:snr_rel} and Table~~\ref{tab:snr_comp}, showing reliability and completeness, split by SNR value, to give a comprehensive view of the performance of our models, and also to make them comparable to traditional source detection algorithms.

\end{appendices}

\newpage

\bibliography{reference}%

\end{document}